\documentclass[10pt,twocolumn,letterpaper]{article}

\usepackage{3dv}
\usepackage{times}
\usepackage{epsfig}
\usepackage{graphicx}
\usepackage{url}
\usepackage{amsmath}
\usepackage{amssymb}
\usepackage{xcolor}
\usepackage{caption}
\usepackage{subcaption}
\usepackage{bm,upgreek}
\usepackage{array}
\usepackage{fixltx2e}
\usepackage{tikz}

\usepackage{pgfplots}
\usetikzlibrary{shadings}
\usetikzlibrary{decorations}
\usetikzlibrary{positioning}

\usepackage[pagebackref=true,breaklinks=true,colorlinks,bookmarks=false]{hyperref}

\threedvfinalcopy 

\def\threedvPaperID{050} 

\ifthreedvfinal\fi
\begin{document}

\title{Model-based Outdoor Performance Capture}

\author{%
Nadia Robertini$^\text{1, 2}$\hspace{0.8em}
Dan Casas$^\text{1}$\hspace{0.8em}
Helge Rhodin$^\text{1}$\hspace{0.8em}
Hans-Peter Seidel$^\text{1}$\hspace{0.8em}
Christian Theobalt$^\text{1}$%
\\\vspace*{1pt}\\
$^\text{1}$ Max-Planck-Institute for Informatics\quad%
$^\text{2}$ Intel Visual Computing Institute%
}

%
%
\twocolumn[{%
	\renewcommand\twocolumn[1][]{#1}%
	\maketitle
	\begin{center}
		\centering%
		\includegraphics[width=\textwidth]{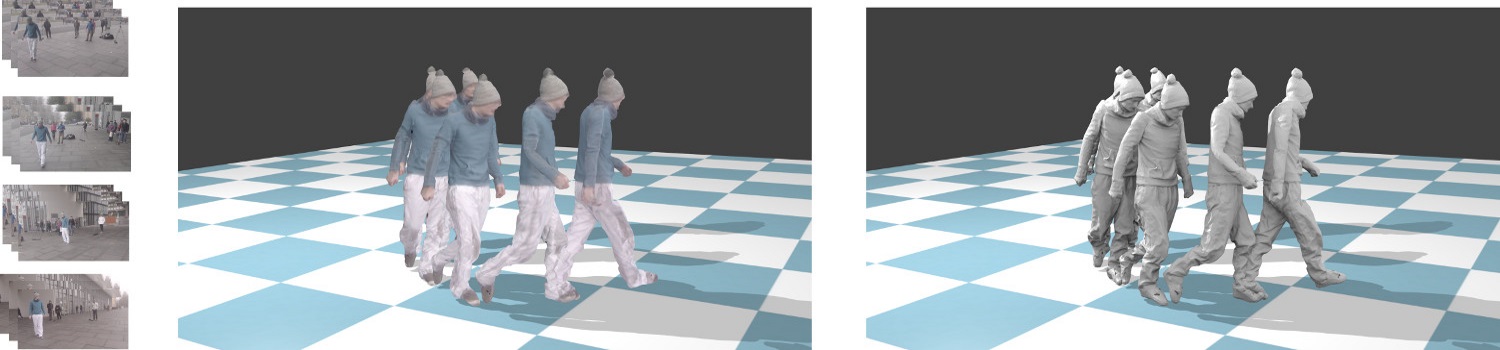}%
		\captionof{figure}{From a set of multi-camera input images (left), we reconstruct the human performance as a temporally consistent 3D mesh that accurately matches the captured motion, here visualized textured (center) and untextured (right) to better aporeciate the results. Our novel formulation for model-based human performance capture enables reconstruction of outdoor performances without explicit silhouette segmentation.}
		\label{fig:teaser}
	\end{center}%
}]

\begin{abstract}
	We propose a new model-based method to accurately reconstruct human performances captured outdoors in a multi-camera setup. 
	Starting from a template of the actor model, we introduce a new unified implicit representation for both, articulated skeleton tracking and non-rigid surface shape refinement.
	Our method fits the template to unsegmented video frames in two stages -- first, the coarse skeletal pose is estimated, and subsequently non-rigid surface shape and body pose are jointly refined.
	Particularly for surface shape refinement we propose a new combination of 3D Gaussians designed to align the projected model with likely silhouette contours without explicit segmentation or edge detection. 
	We obtain reconstructions of much higher quality in outdoor settings than existing methods, and show that we are on par with state-of-the-art methods on indoor scenes for which they were designed. 	
\end{abstract}
%
\section{Introduction}
Marker-less human performance capture methods aim to reconstruct the motion as well as the temporally coherent non-rigid surface geometry of people in their general, potentially loosely deforming, apparel, from multi-view RGB video~\cite{bradley2008markerless, de2008performance, gall2009motion, starck2007surface, vlasic2009dynamic}. 
Several state-of-the-art methods reconstruct highly detailed 3D mesh sequences by fitting a 3D template to the observed performance~\cite{de2008performance,gall2009motion,vlasic2008articulated}; more general methods reconstruct per-frame geometry independently and without a prior model~\cite{franco2009efficient,starck2007surface,vlasic2009dynamic}.
%
%
%
%
Most high-quality reconstruction methods 
fail on footage recorded in general outdoor scenes, as they expect constant lighting and 
crisp foreground/background subtraction which is best achieved in front of static indoor green screens.
%

Aiming to overcome this limitation, recent research in joint segmentation and reconstruction \cite{taneja2010modeling, mustafa2015iccv, mustafa2016cvpr} successfully reconstructed deforming objects in less constrained setups. 
However, resulting 3D mesh detail is significantly lower than for previous in-studio silhouette-based methods. 
Orthogonal to performance capture, marker-less motion capture methods exist which work without silhouettes and in less controlled scenes, but they only estimate coarse human skeleton pose \cite{stoll2011iccv,elhayek2015efficient,amin2013multi,urtasun2006temporal}.
%
%

We propose a new model-based performance capture method that takes a leap forward, and captures detailed human performance, including accurate motion and loose non-rigid surface shape, in less controlled and outdoor environments with moving background, \textit{without} explicit silhouette extraction. 
%
%
%
%
%
To meet the challenges of less controlled environments, we
use a new unified implicit formulation for both, articulated skeleton tracking and non-rigid surface shape refinement. 
Our method fits an initial static surface mesh of an actor to unsegmented video frames in two stages -- first skeletal pose is optimized, and subsequently non-rigid surface shape is refined.
%
%
%
In both stages, we use a mathematical formulation as the minimization of an objective function that estimates the agreement of model and observation.
%

%
The coarse volumetric 3D body shape and body appearance used for skeletal pose estimation, as well as the fine-scale 3D surface geometry and appearance, which is coupled to the coarse body representation, are modeled using the same building block 
-- an implicit function \cite{ilic2006implicit} defined over sets of Gaussians.
Particularly for surface shape refinement we introduce a new combination 3D Gaussians (which we refer to as \textit{Border Gaussians}), designed to align the projected model with likely silhouette contours without explicit segmentation or edge detection.
Also each input image is transformed to a similar implicit representation. 
This scene representation enables effective and efficient pose optimization, and analytically differentiable smooth objective functions for both coarse fitting and refinement on unsegmented images. 
%
%
%

%
Ours is the first integrated template representation and fitting approach for both coarse and fine geometry capable of accurately reconstructing the articulated motion and deforming surface geometry of actors in challenging outdoor scenes.
%
We obtain reconstructions of much higher quality in outdoor settings than existing methods, and show that we are on par with state-of-the-art methods on indoor scenes for which they were designed. 

\section{Related Work}
%
%

\paragraph{Markerless motion capture.}
Marker-less 3D skeletal(-only) motion capture from multi-view video has been extensively studied in the past~\cite{moeslund2012}.
\textit{Generative} methods use actor models composed of primitives, such as ellipses, cylinders and cardboards \cite{wren1997pfinder, bregler1998tracking,ju1996cardboard}, general mesh representations~\cite{ballan2008marker,gall2009motion}, or parametric human models~\cite{sigal2007combined,loper2015smpl}. They then optimize the pose that best explains the observations. 
Our approach is also inspired by methods using implicit surfaces \cite{plankers2003articulated} and volumetric models \cite{stoll2011iccv,rhodin2015iccv}.
\emph{Discriminative} approaches do pose estimation based on trained classifiers \cite{sigal2012loose,amin2013multi,belagiannis20143Dpictorial}; they often play out their power in combination with generative approaches 
 \cite{sigal2007combined,sridhar2015cvpr,elhayek2015efficient}.
While early methods were constrained to controlled indoor settings, more recent methods succeeded on outdoor scenes, even 
when recorded with a low number of cameras~\cite{elhayek2015efficient,rhodin2015iccv}

We use a variant of Stoll \etal.~\cite{stoll2011iccv} for coarse pose estimation. Generally motion capture methods are complemented with refinement approaches as they retain no surface detail and are inaccurate in situations of strong surface deformation, such as loose clothing forming folds. Nevertheless, they reconstruct the coarse human pose reliably. 
%


\paragraph{Performance capture.}
Existing methods for performance capture can be split into two trends.
On one hand, \emph{model-based} methods \cite{de2008performance,gall2009motion,vlasic2008articulated,liu2011markerless,wu2013set,wu2012full} deform a static template of an actor, usually acquired with a full body scanner,
 to best fit it into synchronized multi-camera input.
Parametric body models \cite{sminchisescu2003estimating,jain2010movie,anguelov2005scape,pishchulin2015building,loper2014mosh,loper2015smpl,bogo2015detailed,rhodin2016eccv} can also be used as a geometric template, however they are not well suited to represent surface deformations caused by cloth.
%
On the other hand, \emph{model-free} methods \cite{starck2007surface,vlasic2009dynamic,franco2009efficient,collet2015high} remove the need of an initial template by reconstructing a per-frame independent geometry --- using visual hull on silhouettes \cite{franco2009efficient} and multi-view photometric stereo reconstruction techniques \cite{starck2007surface}. 
Note that, in contrast to model-based methods, the output geometry is temporally incoherent (\eg different amount of vertices and edges per frame), and needs to be later temporally tracked and aligned \cite{budd2013global,cagniart2010probabilistic} to achieve a compact temporally coherent 4D representation.

A key limitation of many existing performance capture methods is their dependency on explicit background segmentation for accurate silhouette alignment, an error-prone step which hinders their usage in uncontrolled environments.
%
Progress has been made by multi-view segmentation \cite{wang2014wide,djelouah2015sparse}, joint segmentation and reconstruction \cite{szeliski1998stereo,guillemaut2011joint,bray2006simultaneous,mustafa2015iccv,mustafa2016cvpr,cui2012kinectAva}, and also aided by propagation of a manual initialization \cite{hasler2009cvpr,wu2013set,taneja2010modeling}.
In uncontrolled environments the obtained segmentation is still noisy, enabling only skeleton pose \cite{hasler2009cvpr}
and rather coarse 3D reconstructions~\cite{mustafa2015iccv,mustafa2016cvpr}.
Rhodin et al.~propose a volumetric contour model and directly fit a parametric shape model to image edges, circumventing silhouette extraction entirely. However, only coarse shape without cloth-level detail is reconstructed \cite{rhodin2016eccv}.

Surface refinement in performance capture is particularly relevant to our work. Some methods deform an initial mesh fit by pulling the surface vertices towards the silhouette contours \cite{vlasic2008articulated,de2008performance}.
More sophisticated methods use inverse rendering techniques that refine coarse geometry using shading cues \cite{wu2011shading,wu2013set}.
Most related to our work, recently Robertini \etal~\cite{robertini20143dv} efficiently recovered medium-frequency surface details optimizing the model-to-image photo-consistency using implicit representations of mesh and images.
Similarly, Ilic \etal \cite{ilic2006implicit} also leverage the attractive properties of implicit surfaces --- \eg differentiability --- for surface reconstruction.
We also use an implicit representation, but tailored for efficient model-to-image edge similarity.

Despite all recent progress in performance capture and surface refinement, existing model-based methods do not cope well in uncontrolled outdoor scenes. Our goal is to enable more accurate model-to-image fitting in challenging scenes.

\begin{figure*}
	\includegraphics[width=\linewidth]{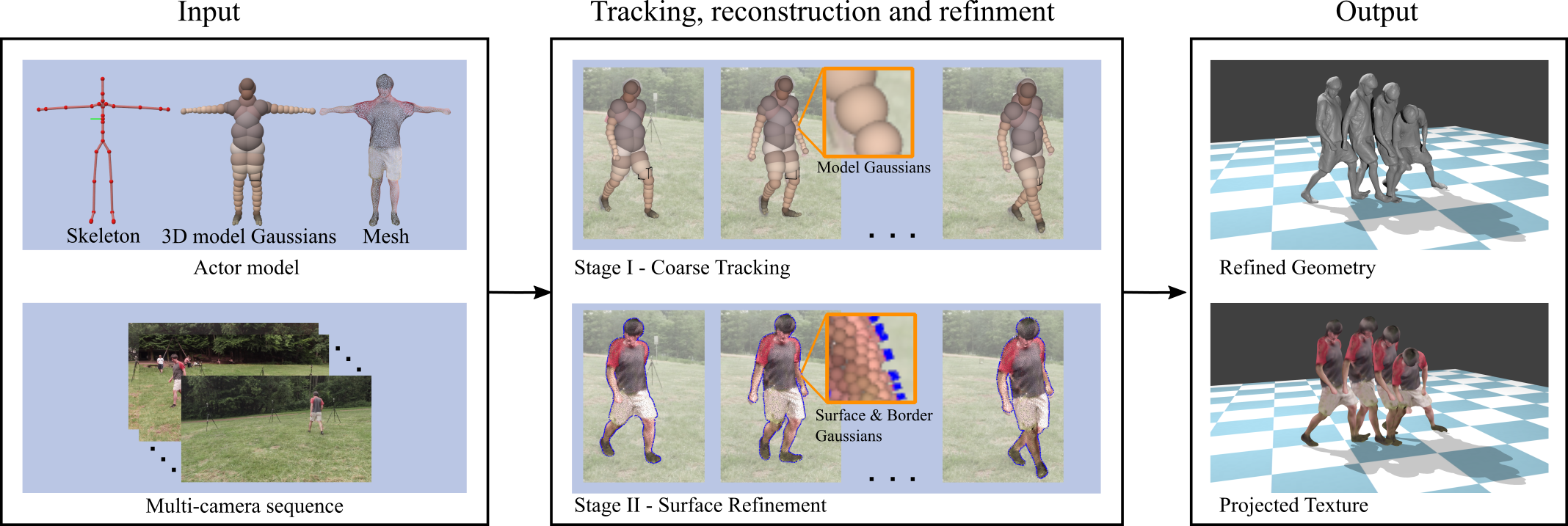}%
	\caption{\label{fig:overview}%
		Overview of our method. Input to our optimization approach is an actor model and a multi-view sequence obtained from synchronized and calibrated cameras (right). The actor model consists of a skeleton with a colored implicit Gaussian-based volumetric representation of the actor as well as a colored static 3D mesh. We optimize model-image agreement in two stages (depicted in the middle), where we subsequently estimate the skeleton pose and then refine the surface using a new tracking approach that approximates the input shape with a set of Gaussians, as explained in Section \ref{sec:model_representation}. Output of our method (left) is a sequence of refined geometry and texture, which best resembles the input performance in terms of pose and surface details.
	}
	\label{fig:overview}
\end{figure*}%
%



\section{Overview}
%
%
%
Our goal is to deform our template model such that it accurately reproduces the performance filmed with calibrated and synchronized multi-view video (Figure \ref{fig:overview}).
To build our actor model we expect as input a colored static 3D mesh of the person obtained through a 3D laser scan, as in~\cite{de2008performance}, through parametric model fitting \cite{balan2007detailed, loper2015smpl, rhodin2016eccv}, or through image-based reconstruction from manually created single-time step silhouettes~\cite{starck2007surface, franco2009efficient}. Then, a kinematic skeleton is fitted to this surface semi-automatically using linear-blend skinning. 

We optimize the model-to-image agreement in two stages.
%
%
Stage-I (Section \ref{sec:coarseTracking}) tracks
the coarse skeleton motion of the performance, based on the approach from Stoll \etal \cite{stoll2011iccv},  
which does not require segmentation and attains high performance through a Gaussian representation of image and actor model (see Section \ref{sec:model_representation}). 
Output is an intermediate skeleton motion, which is used to drive the template mesh by skinning. It fails in reproducing non-rigid deformations caused by cloth and soft tissue deformation, and suffers from skinning artifacts.
Stage-II estimates high-detail surface shape 
by maximizing the agreement between a fine-scale implicit representation of the surface mesh and the image. The detailed representation is obtained by  placing small 3D {\it Surface Gaussians} on each vertex of the mesh and by additionally introducing special \textit{Border Gaussians} used for contour alignment without background segmentation (see Section \ref{sec:surfaceRefinement}). 
While refining the surface, also the skeleton obtained in Stage-I is jointly refined. 

The output of our method is a sequence of skeletal poses alongside with refined surface meshes that reproduce the input videos. 
%

\section{Performance Capture}
\subsection{Model Representation}
\label{sec:model_representation}
%
We use a two-layer representation to express our actor model. \textit{Layer-I} consists of
a skeleton $\mathcal{S}=\{\bm{\uptheta}, \mathbf{\hat{g}}\}$, where $\bm{\uptheta} = \{\theta_j\}_{j=1}^J$ is a degree of freedom vector that parameterizes the skeleton pose by its joint angles, and $\mathbf{\hat{g}} = \{\hat{g}_m\}_{m=1}^M $ a collection of colored 3D Gaussians (\textit{Model Gaussians}), rigidly attached to the skeleton, that approximate the 3D volume of the actor coarsely (see Figure \ref{fig:overview} top-left and \cite{stoll2011iccv}).
%
%
The skeleton is once fitted to an actor mesh $\mathcal{M}$ during preprocessing, and skinning weights are computed in a static pose. Each Model Gaussian is assigned the average color of the surrounding mesh vertices.

\textit{Layer-II} of the model is derived from the rigged, colored static surface mesh $\mathcal{M} = \{ \mathbf{v}, \mathbf{W} \}$, where $\mathbf{v} = \{ v_n \}_{n=1}^N$ are the vertices, and $\mathbf{W} \in \mathbb{R}^{J \times N}$ a matrix of rigging weights that define how each vertex $v_n$ moves with respect to the degree of freedom $\theta_j$.
We do not explicitly use the mesh for capturing, but transform its vertices into an implicit representation $\mathcal{G} = \{\mathbf{\tilde{g}}^c, \mathbf{\breve{g}}^c\}$ consisting of two subsets of Gaussians.
The first set are \textit{Surface Gaussians} $\mathbf{\tilde{g}}^c =\{\tilde{g}_{s}\}_{s=1}^{S_c}$ for the visible vertices from camera $c$ that do not fall onto an occluding mesh contour at a given frame.
A small 3D Gaussian is placed at each vertex position $v_n$ and is assigned the color of the static mesh vertex \cite{robertini20143dv}.
The second set are \textit{Border Gaussians}  $\mathbf{\breve{g}}^c =\{\breve{g}_b\}_{b=1}^{B_c}$ for the vertices that lie on an outer occluding mesh contour in the original camera view $c$.
%
%
Note that on {\it Layer-II}, we preserve the mesh connectivity information to define the surface topology of Surface and Border Gaussians, and assume they are coupled to the skeleton by the original mesh skinning weights. 
%
%
%
%
%
See Figure~\ref{fig:overview} top-left for a visualization of the actor model. 

Additionally, our input image set $\mathcal{F} = \{ \mathbf{f}^c \}_{c=1}^C$, where $C$ is the number of cameras, is approximated with a set of 2D Gaussians (\textit{Image Gaussians}) $\mathcal{I} = \{ \mathbf{i}^c \}_{c=1}^C$, where $ \mathbf{i}^c$ is a vector of 2D Gaussians that approximate the image frame $ \mathbf{f}^c$.
The na\"ive approach is to create a single Image Gaussian for each pixel, and assign to it the pixel color. However, this results in an excessive amount of elements.
Instead, as in \cite{stoll2011iccv}, we use a quad-tree decomposition to efficiently cluster each frame with similar color areas into a single Gaussian. 
%


\subsection{Stage-I -- Coarse Tracking}
\label{sec:coarseTracking}
In this stage an initial skeletal pose $\bm{\uptheta}$ is estimated using the Layer-I model and the algorithm from~\cite{stoll2011iccv}, where we use the commercial implementation 
available through The Captury Studio \cite{TheCaptury}.
This approach optimizes the skeleton such that the attached Model Gaussian $\mathbf{\hat{g}}$, projected on the image, agrees in color with the nearby Image Gaussians ${\mathbf{i}_c}$, see Figure \ref{fig:overview} top-center. 
The core of the method is a pairwise Gaussian overlap measure, which estimates the proximity of Model and Image Gaussians,
%
\begin{align}
E_{m,i} &=  
\left[\int_{\Omega} \hat{g}_m(x) \hat{i}_c(x) \partial x \right]^2
 = 
2 \frac{\sigma_m\sigma_{c}}{\sigma_{m}^2+\sigma_{c}^2} e^{-\frac{||\mu_m - \mu_{c} ||^2}{\sigma_{m}^2+\sigma_{c}^2}},
\label{eqn:GaussianOverlapPair}
\end{align}
where $\mu_i,\sigma_i$ are the Image Gaussian mean and standard deviation, respectively, $\mu_m$ is the Model Gaussian mean projected on the image plane by scaled orthographic projection, and $\sigma_m$  the projected Model Gaussian size.


While any other outdoor motion capture algorithm could be used instead, the analytic and smooth form of the Gaussian overlap integral 
 is of interest for our surface refinement method, which is explained in the next section.
\subsection{Stage-II -- Surface Refinement}
\label{sec:surfaceRefinement}
%
Stage-I only captures the coarse articulated pose, but no fine-scale non-rigid surface deformations are well recovered by just rigging the template $\mathcal{M}$ using the joint angles $\bm{\uptheta}$. 
We therefore compute a refined estimate using the Layer-II model by finding $(\mathbf{v},\bm{\uptheta})$ maximizing
%
\begin{equation}
	\begin{aligned}
		E(\mathbf{v},\bm{\uptheta}) 
		=\ & 
		E_{\text{surf}}(\mathbf{v}) 
		+ 
		E_{\text{cont}}(\mathbf{v})\ - \\
		&w_{\text{skin}}E_{\text{skin}}(\mathbf{v},\bm{\uptheta})
		- 
		w_{\text{smooth}}E_{\text{smooth}}(\mathbf{v}),
	\end{aligned}
	\label{eq:StageII}
\end{equation}
%
%
%
initialized with the pose $\bm{\uptheta}$ and the associated skinned mesh $\mathbf{v}$. In the following we explain the individual energy terms.

%
%
{$E_{\text{surf}}$} measures the photo consistency of the \textit{visible} mesh surface vertices with the input images.
It is implemented as a generalization of the volumetric Gaussian tracking described in Stage-I to represent the mesh surface implicitly.
%
%
\begin{align}
E_\text{surf}(\mathbf{v}) = \sum_s^{|\mathbf{\tilde{g}}|} \sum_i^{|\mathbf{i}|} C(\delta_{s,i}) E_{s,i},
\label{eqn:GaussianOverlapSum}
\end{align}
where $|\mathbf{\tilde{g}}|$ is the number of Surface Gaussian, and $E_{s,i}$ is the Gaussian overlap between Surface and Image Gaussians, defined in Equation \ref{eqn:GaussianOverlapPair}. 
The color similarity $C(\delta_{s,i})$, which maps the HSV color difference $\delta_{s,i}$ to the range $[0,1]$ with a smooth step function, is a robust measure of Surface and Image Gaussian.
%
This term is similar to \cite{robertini20143dv} and accounts for fine detail refinement of the surface interior, when texture cues are available, but does not account for accurate contour alignment.

{$E_{\text{cont}}$} measures the model-to-image contour alignment.
%
%
Our goal is to align each border vertex in color, space and direction with nearby image gradients, \ie move border vertices such that their projection: (1) spatially coincides with a strong edge, (2) shows a strong gradient from vertex color to background color, and (3) the edge orientation aligns with the mesh contour direction.
In our setting, we have an accurate shape and appearance model of the actor, but face unknown background, \eg~moving scenes, which hinders direct foreground-background gradient computation.

\begin{figure}
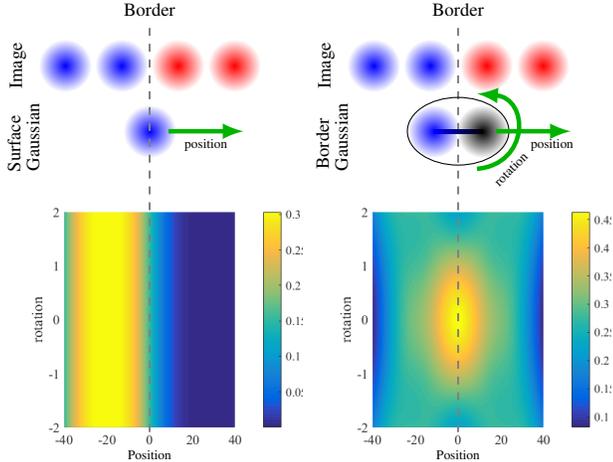

	\begin{subfigure}[t]{0.45\linewidth}
		\input{figures/single_blob_energy_landscape.tikz}
	\end{subfigure}
~~
	\begin{subfigure}[t]{0.45\linewidth} 
		\input{figures/pair_blob_energy_landscape.tikz} 
	\end{subfigure}
	\caption{Energy landscape for a toy example using: (a) Surface Gaussian; and (b) Border Gaussian. From top to bottom, in each subfigure: 2D Gaussian test image, Gaussian sampled along the image, and energy landscape.}	
	\label{fig:GaussianOverlap}
\end{figure}

We propose a formulation that neither requires pre-computations nor knowledge of the background color and is nevertheless efficient to optimize.
For each mesh vertex $v_n$ that is within a $\Delta$-distance to the contour of the mesh in the camera plane, we create an implicit representation, which we referred to as \textit{Border Gaussian} $\breve{g}_b$, by placing two 3D Gaussians:
an \emph{Inside Gaussian}, displaced $\sigma_b$ inside along the surface normal with $C(\delta_{b,i})$ as before; and an \emph{Outside Gaussian}, displaced by $\sigma_s$ to the outside and with color similarity $(1 - C(\delta_{b,i}))$.
We therefore optimize
\begin{align}
E_\text{cont}(\mathbf{v}) = \sum_b^{|\mathbf{\breve{g}}|} \sum_i^{|\mathbf{i}|} C(\delta_{b_,i}) E_{b_{\text{in}},i} + (1-C(\delta_{b,i})) E_{b_{\text{out}},i}
\label{eqn:GaussianOverlapSum}
\end{align}
where $|\mathbf{\breve{g}}|$ is the number of Border Gaussian, and $E_{b_{\text{in}},i}$ and $E_{b_{\text{out}},i}$ represent the Gaussian overlap (Equation \ref{eqn:GaussianOverlapPair}) between the Inside and Outside 3D Gaussians and the Image Gaussian, respectively. This optimization causes attraction of the Inside Gaussian to the model color and the Outside Gaussian to the background color, because $(1 - C(\delta_{s,i}))$ is large when the color is dissimilar to the foreground, see Figure \ref{fig:GaussianOverlap}.
This pair of Gaussians approximates a gradient from foreground to background color and has maximal response when the desired alignment of color, space and direction is maximal.

$E_\text{skin}$ couples skeleton and surface motion. Some parts of the body are mostly rigid, such as shoes, and require less surface deformation refinement, while other parts, such as clothing, are more deformable.
To model this we link the set of Surface Gaussian to the rigid transformations $T_j$ of each skeleton joint $j$ by minimizing the Surface Gaussian distance to the skinned position $\check{\mu}_s$, 
\begin{equation}
\check{\mu}_s = \sum_{j=1}^{J} W(j,n) T_j d_{j,n}, 
\end{equation}
where $W(j,n)$ is the skinning weight of vertex $n$ by joint $j$, and $d_{j,n}$ is the rigid offset between the vertex and the joint. 
The distance between refined and skinned position is estimated using the 3D overlap of the optimized Surface Gaussian $\hat{g}_s$ with its rigid correspondent $\check{{g}}_s$, using Equation~\ref{eqn:GaussianOverlapPair}. 
A rigidity weight is defined for each vertex by providing a mask, which enables to regularize different parts of the body differently (\eg allow loose clothing to move \textit{more} freely
than hands or feet).
This also allows as to deal with difficult situations where body parts are partly occluded, as demonstrated in Figure \ref{fig:rigidity_evaluation}, where the shoe is hidden in high grass and the surface refinement step erroneously squeezes the geometry. When using a rigidity mask, we can enforce vertices on the feet to maintain the original $\check{\mu}_s$.
Additionally, as a consequence of jointly optimizing $\mathbf{v}$ and $\bm{\uptheta}$, the $E_{\text{skin}}$ term also refines the skeleton pose. See the supplementary video for a visualization.

\newcommand{\trimmedRiggidTerm}[2][]{%
	\includegraphics[trim={1cm 1cm 1cm 1cm},clip, width=0.174\linewidth,#1]%
	{#2}%
}
\newcommand{\trimmedRiggidTermZoom}[2][]{%
	\includegraphics[trim={2cm 1cm 6cm 11.3cm},clip, width=0.197\linewidth,#1]%
	{#2}%
}
\begin{figure}
	\begin{tikzpicture}[image/.style={inner sep=0pt},]
	\node[image,right=0pt] (F0) at (0,0)		
	{\trimmedRiggidTerm{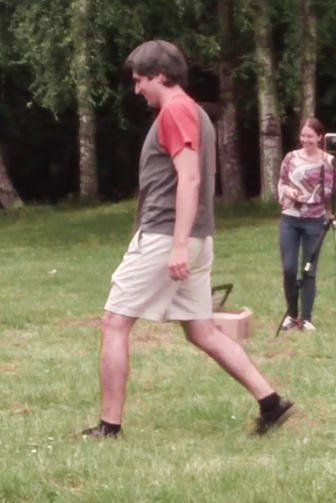}};
	\node[scale=0.8,below=2pt] at (F0.south) {Input};
	\draw[red, thick,densely dashed] (0.125,-0.25) rectangle (0.7,-1.1);
	\node[image,right=2pt] (F1) at (F0.east)		
	{\trimmedRiggidTermZoom{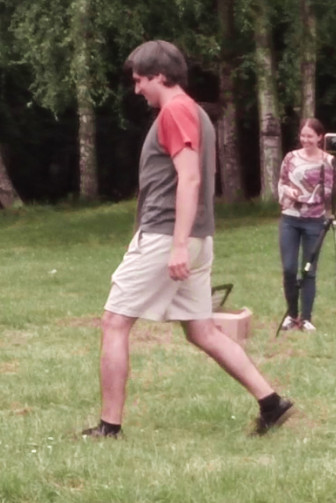}};
	\node[scale=0.8,below=2pt] at (F1.south) {Zoom in};
	\node[image,right=2pt] (F2) at (F1.east)		
	{\trimmedRiggidTermZoom{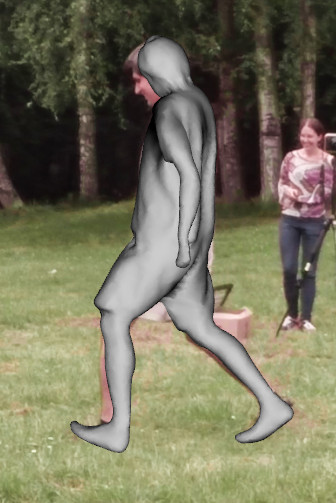}};
	\node[scale=0.8,below=2pt] at (F2.south) {Stage-I};
	\node[image,right=2pt] (F3) at (F2.east) 
	{\trimmedRiggidTermZoom{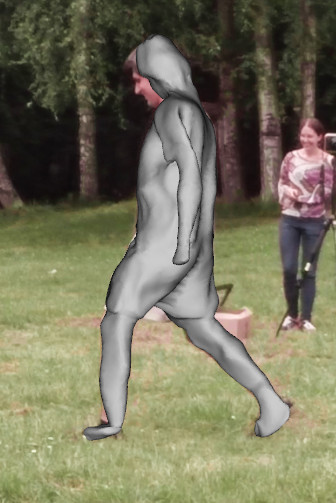}};
	\node[scale=0.8,below=2pt] (F3_label) at (F3.south) {Stage-II, no};
	\node[scale=0.8,below=-3pt]  at (F3_label.south) {rigidity mask};
	\node[image,right=2pt] (F4) at (F3.east)
	{\trimmedRiggidTermZoom{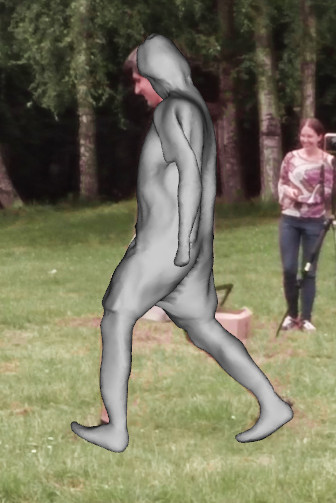}};
	\node[scale=0.8,below=2pt] (F4_label) at (F4.south) {Stage-II, with};
	\node[scale=0.8,below=-3pt]  at (F4_label.south) {rigidity mask};
	\end{tikzpicture}
	\caption {Influence of the rigidity mask on the $E_{\text{skin}}$ component. In this challenging situation where the foot is occluded by the high grass, our refinement step in Stage-II, strongly influenced by the data term $E_{\text{surf}}$, tends to implausibly squash the geometry to maximize model-to-observation similarity. A rigidity mask can be used to define the local rigidness of each vertex. On the right, notice how the lower leg is refined such that it matches the image contour, while the foot maintains its original volume.}
	\label{fig:rigidity_evaluation}
\end{figure}
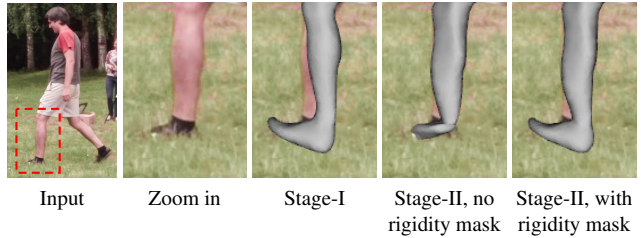

$E_{\text{smooth}}$ regularizes unnatural surface deformations with a smoothness prior term. 
We use a Laplacian smoothness term which peanalizes deviations of the optimized mesh Laplacian 
from the mesh Laplacian of the template mesh, skinned by the initial $\bm{\uptheta}$. 
Note that \cite{robertini20143dv} constrained vertex motion explicitly to the surface normal direction, which suppresses tangential corrections entirely and can preclude convergence from large displacements, while the Laplacian regularization only ensures coherent motion of nearby vertices.


The unified energy representation \label{eq:StageII} enables joint and efficient optimization of surface interior, contour alignment, and skeleton pose, 
using traditional gradient ascent.
In Section \ref{sec:results_and_evaluation}, we show that the proposed model applies to more general scenes than existing model based approaches and that exploiting of a detailed actor model gives superior results to existing model-free methods. 
%
\definecolor{MyPurple}{rgb}{0.63,0.63,1}
\section{Results and Evaluation}
\label{sec:results_and_evaluation}
We qualitatively and quantitatively evaluate our method, and compare our results with state-of-the-art human performance capture methods, both indoor and outdoors.
See the supplementary video for further evaluation and results.
The supplementary document contains a description of each sequence and actor model.

All results presented in this section were computed on a desktop computer with a NVIDIA GeForce Titan X.
Using unoptimized code, our Stage-II run time is about 5--30 minutes on the CPU and 1--4 minutes on the GPU, per frame.

\subsection{Quantitative Evaluation}
\label{sec:quantitative_evaluation}
To quantitatively assess the performance of our method, we use a silhouette overlap metric between the actor model projected to the camera plane and the ground truth silhouette obtained with manual segmentation.
We label as \textit{positive} and \textit{negative} the foreground and background pixels, respectively. When the label of the projected model and the ground truth image pixel agree, it is a \textit{true} pixel, and \textit{false} otherwise.
The combination of true and false pixels is expressed with the F\textsubscript{1} score, commonly used in statistical analysis for binary classification, which can be interpreted as a weighted average of \textit{precision} and \textit{recall}. 
Note that figures in this section visualize the resulting overlap labels using \textcolor{green}{\textbf{Green}} for false negative, \textcolor{red}{\textbf{Red}} for false positive, \textcolor{MyPurple}{\textbf{Purple}} for true positive and \textbf{Black} for true negative.

Figure \ref{fig:skirtEvaluation} presents a visualization of the qualitative evaluation of the \texttt{skirt} sequence --- an indoor sequence, with ground truth silhouettes available --- showing the overlap of the mesh and the ground truth contour (top row), and the resulting silhouette overlap labels (middle row), in 4 different settings.
As expected, the mesh generated in Stage-I (first column) is incapable of capturing the non-rigid skirt contour and suffers from skinning artifacts in the shoulder area.
Stage-II (second column) significantly improve these shortcomings, resulting in a much accurate alignment.
To further evaluate our approach, we enforce Stage-II to work in ideal conditions where the background is known (third column).
Instead of working with the input color images, we use the silhouette images --- \ie background is black --- and therefore we assign the Outside Gaussians of the Border Gaussian also a black color, instead of the inverse of the inner Gaussian color as we do in uncontrolled conditions. 
Results under such ideal conditions are shown in the third column, and further validate our new implicit representation: \textit{perfect} color assignment of the inner and Outside Gaussians refines the mesh such that it perfectly matches the ground truth.
Our results are in fact comparable to Gall \etal \cite{gall2009motion} (fourth column), a state-of-the-art method that requires explicit silhouette segmentation.

\newcommand{\trimmedSkirtEvaluation}[2][]{%
	\includegraphics[trim={6cm 7cm 11cm 5.5cm},clip, width=0.24\linewidth,#1]%
	{#2}%
}
\newcommand{\trimmedSkirtEvaluationZoom}[2][]{%
	\includegraphics[trim={9cm 9.7cm 16cm 13cm},clip, width=0.12\linewidth,#1]%
	{#2}%
}
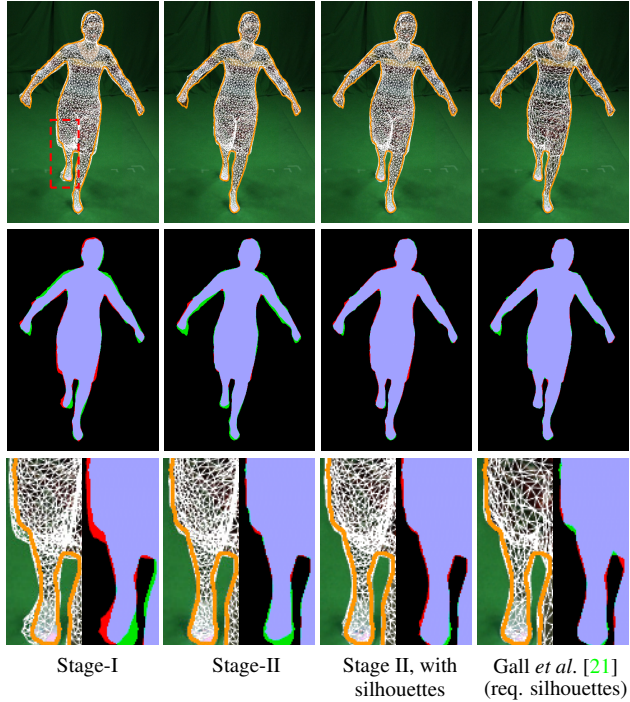
\begin{figure}
	\begin{tikzpicture}[image/.style={inner sep=0pt},]
\node[image,right=0pt] (F0) at (0,0)		
{\trimmedSkirtEvaluation{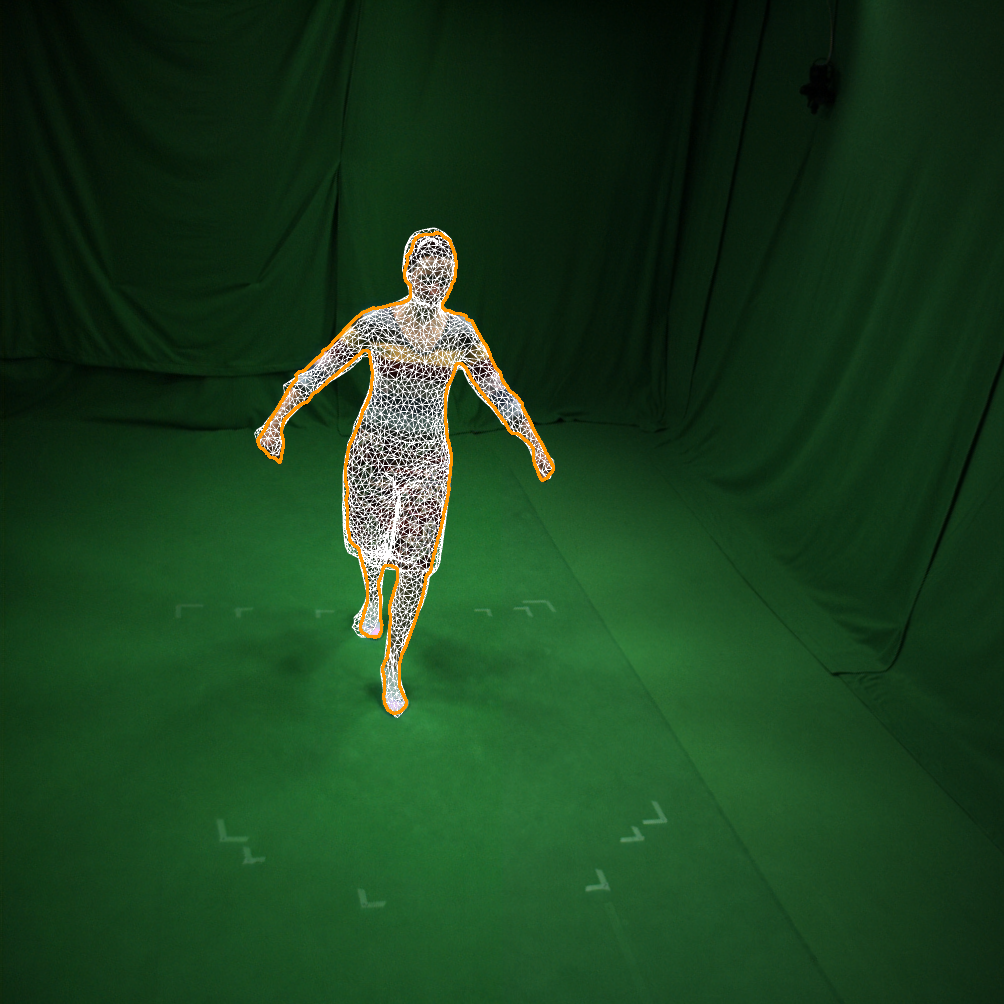}};
\node[image,right=2pt] (F1) at (F0.east)		
{\trimmedSkirtEvaluation{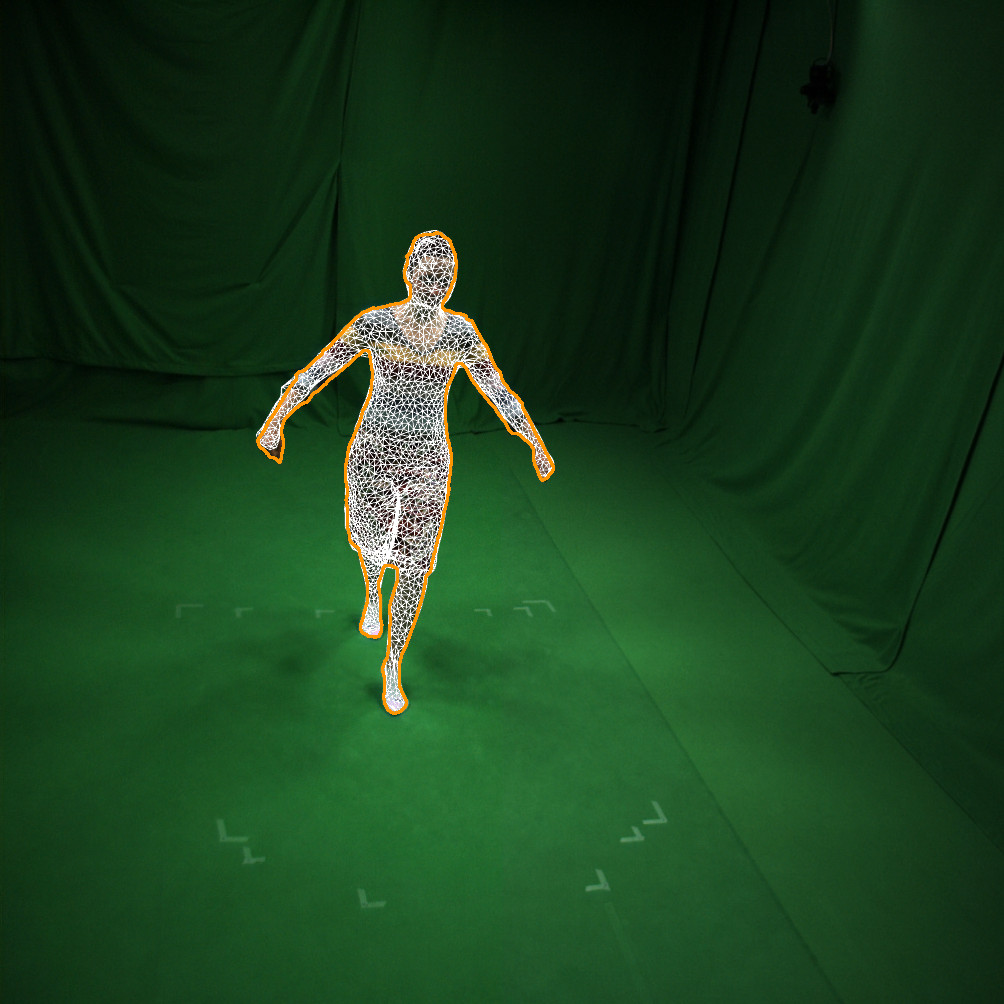}};
\node[image,right=2pt] (F2) at (F1.east)		
{\trimmedSkirtEvaluation{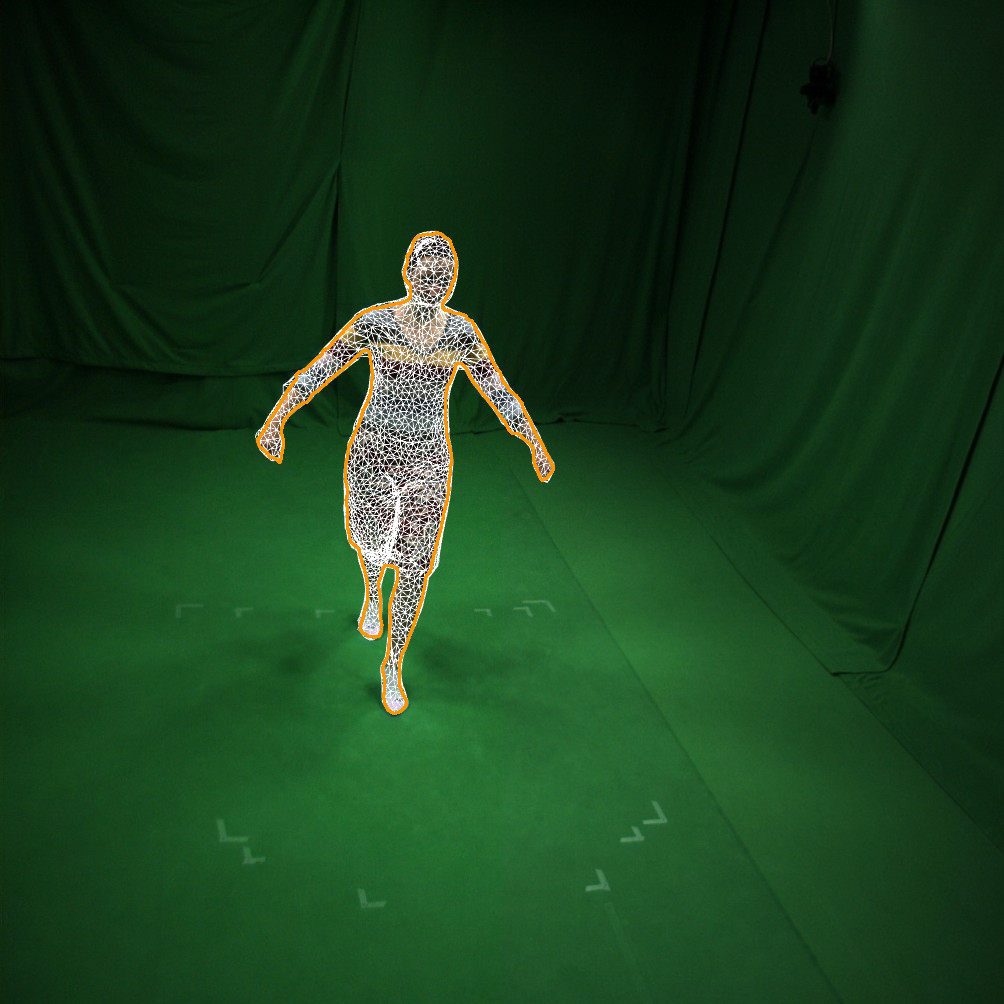}};
\node[image,right=2pt] (F3) at (F2.east)		
{\trimmedSkirtEvaluation{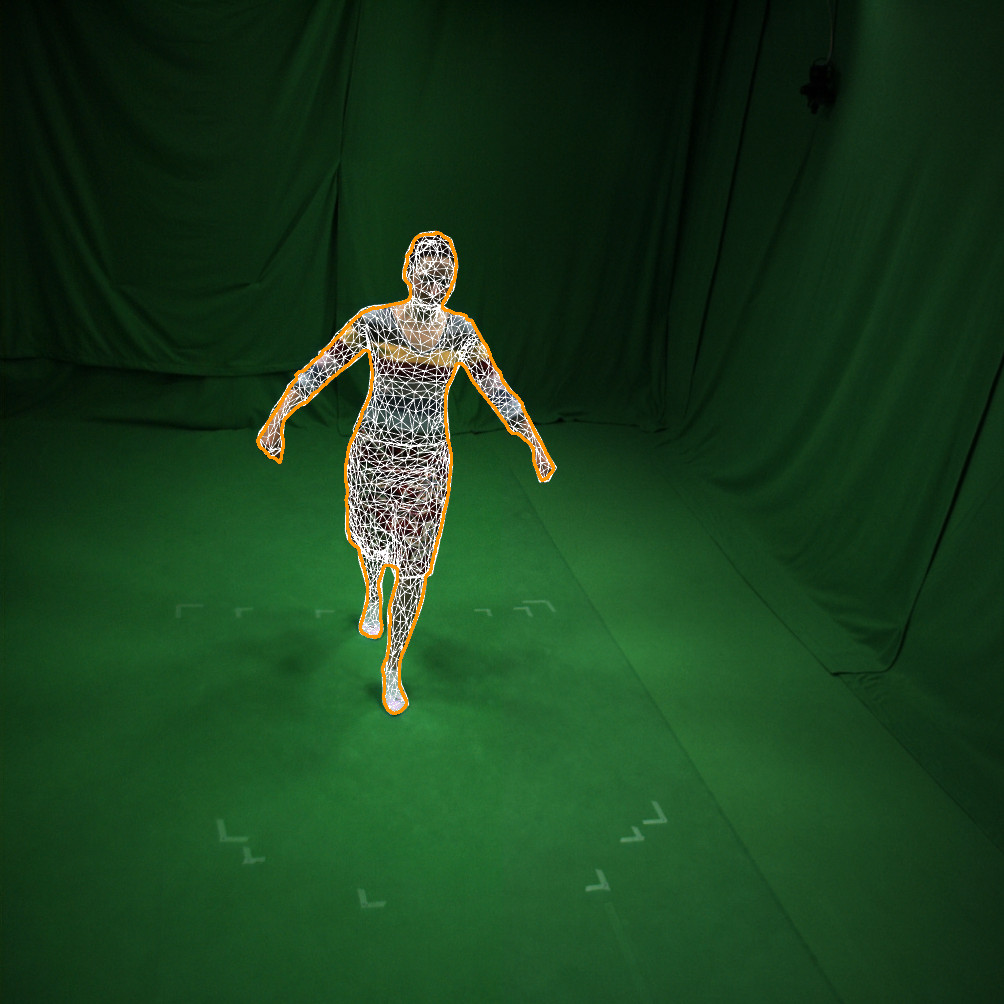}};
\node[image,below=2pt] (F4) at (F0.south)		
{\trimmedSkirtEvaluation{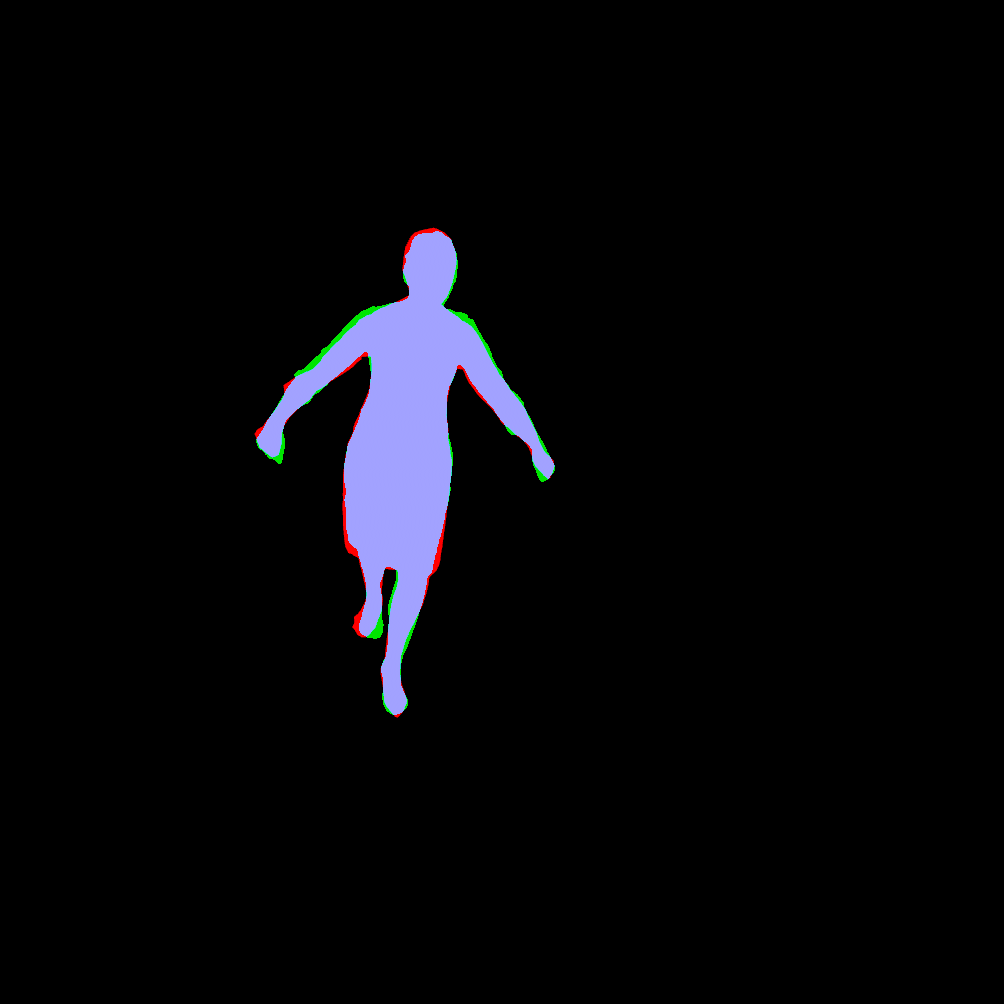}};
\node[image,below=2pt] (F5) at (F1.south)		
{\trimmedSkirtEvaluation{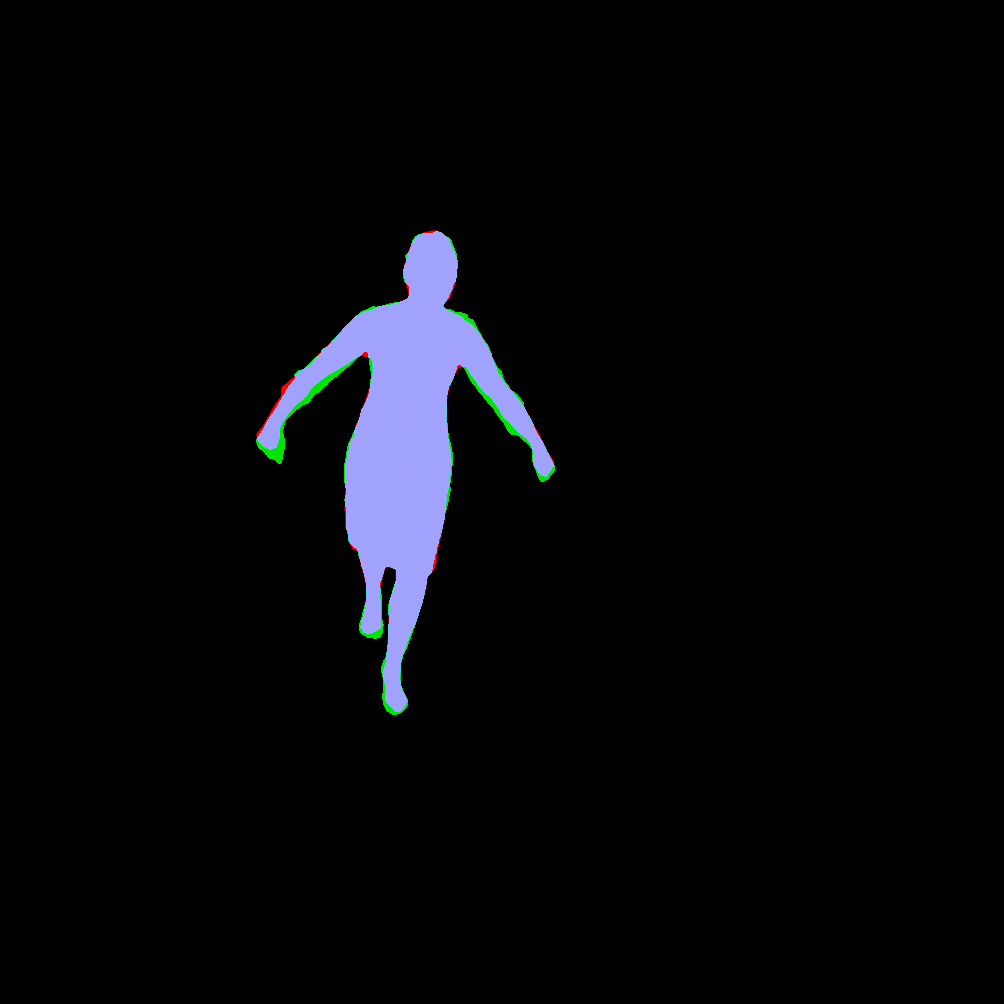}};
\node[image,below=2pt] (F6) at (F2.south)		
{\trimmedSkirtEvaluation{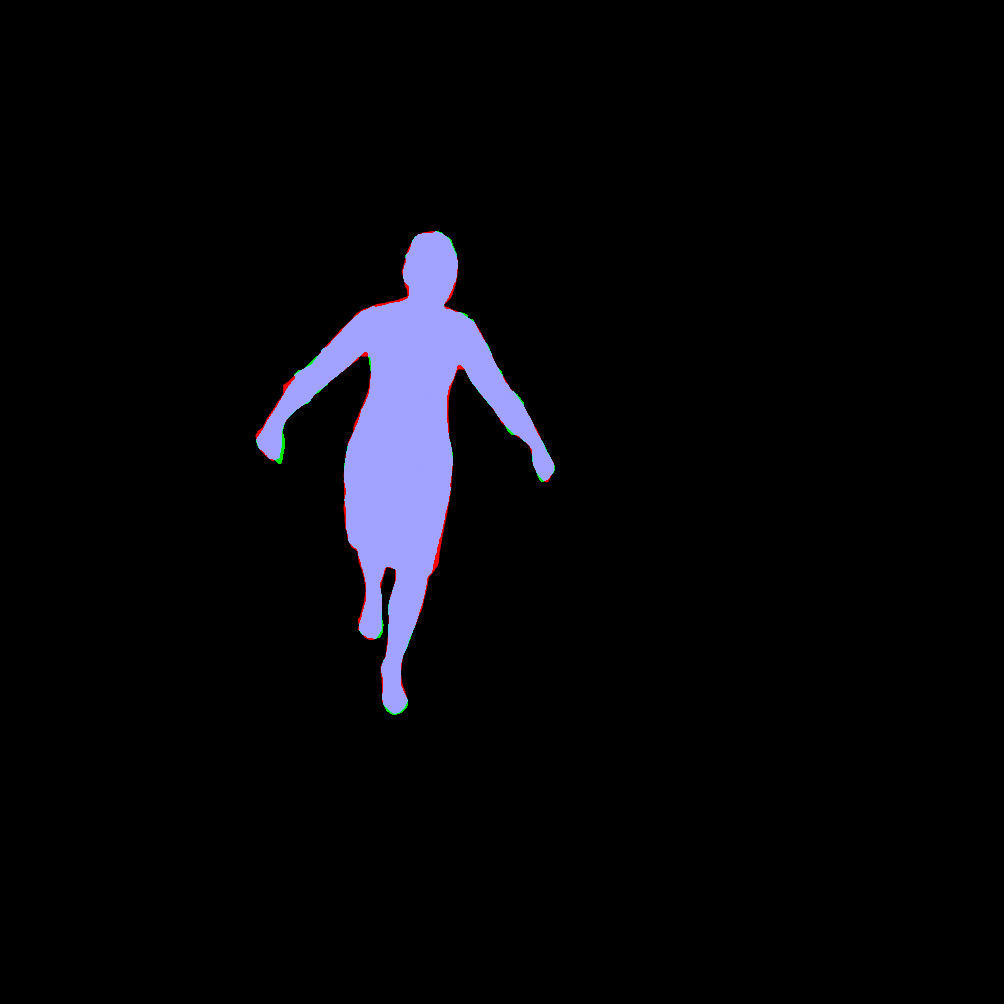}};
\node[image,below=2pt] (F7) at (F3.south)		
{\trimmedSkirtEvaluation{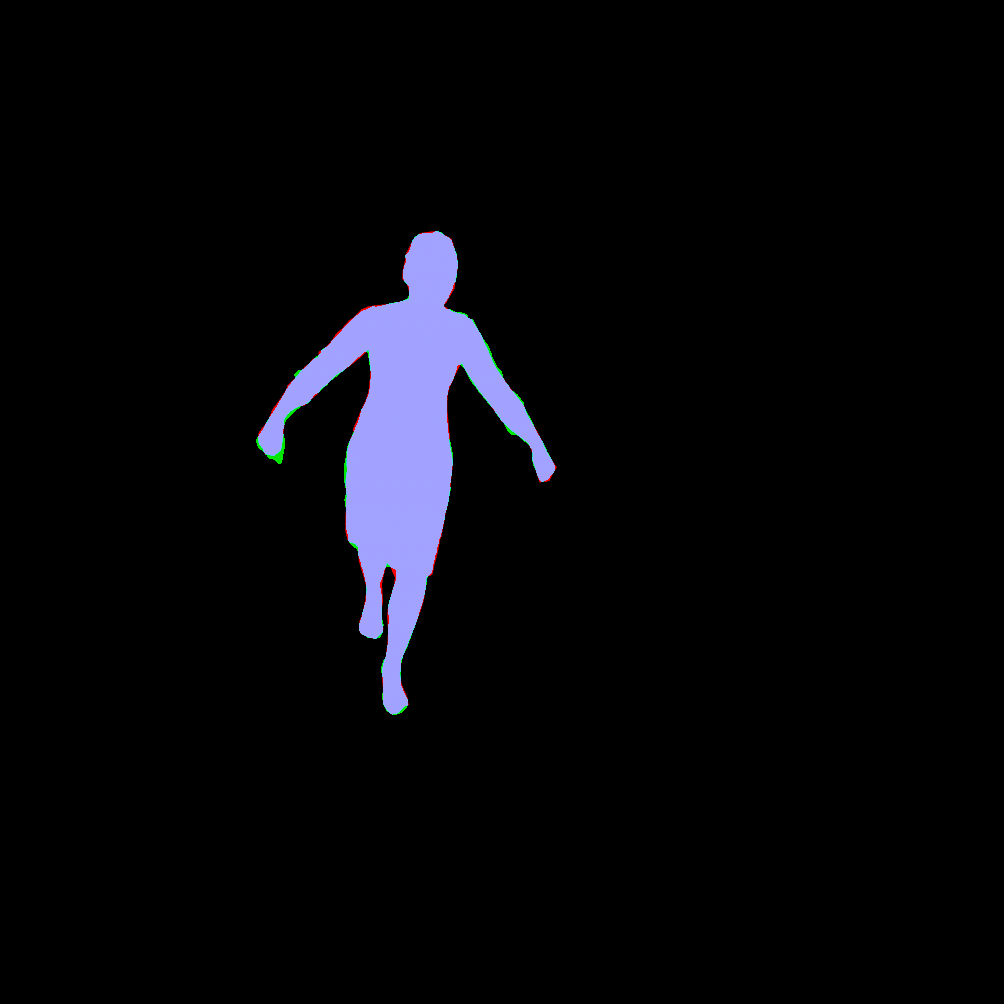}};
\node[image,below left = 0.1cm and -1.0cm of F4] (F8)
{\trimmedSkirtEvaluationZoom{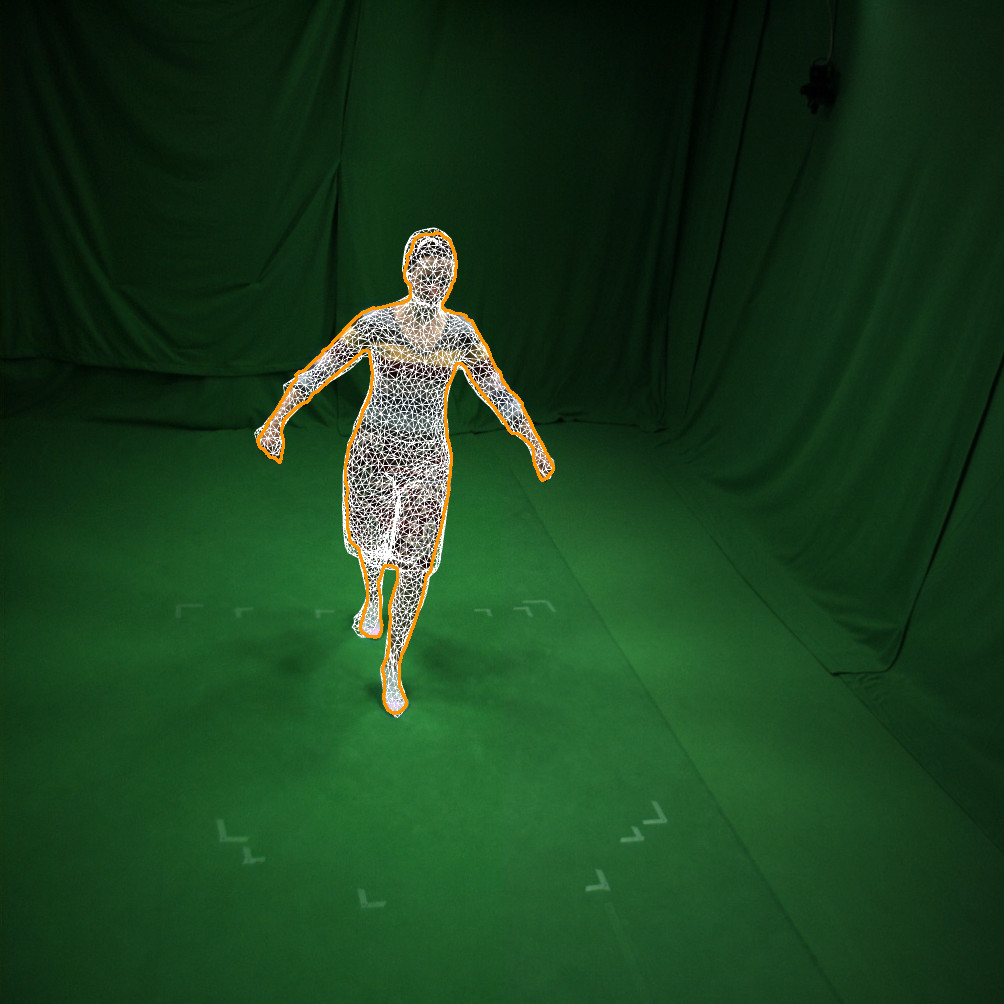}};
\node[image,right=0pt] (F9) at (F8.east)		
{\trimmedSkirtEvaluationZoom{figures/overlap_evaluation/skirt/cam01/baseline_evaluation.png}};
\node[image,below left = 0.1cm and -1.0cm of F5] (F10)
{\trimmedSkirtEvaluationZoom{figures/overlap_evaluation/skirt/cam01/inner_and_outer_overlap.jpg}};
\node[image,right=0pt] (F11) at (F10.east)		
{\trimmedSkirtEvaluationZoom{figures/overlap_evaluation/skirt/cam01/inner_and_outer_evaluation.png}};
\node[image,below left = 0.1cm and -1.0cm of F6] (F12)	
{\trimmedSkirtEvaluationZoom{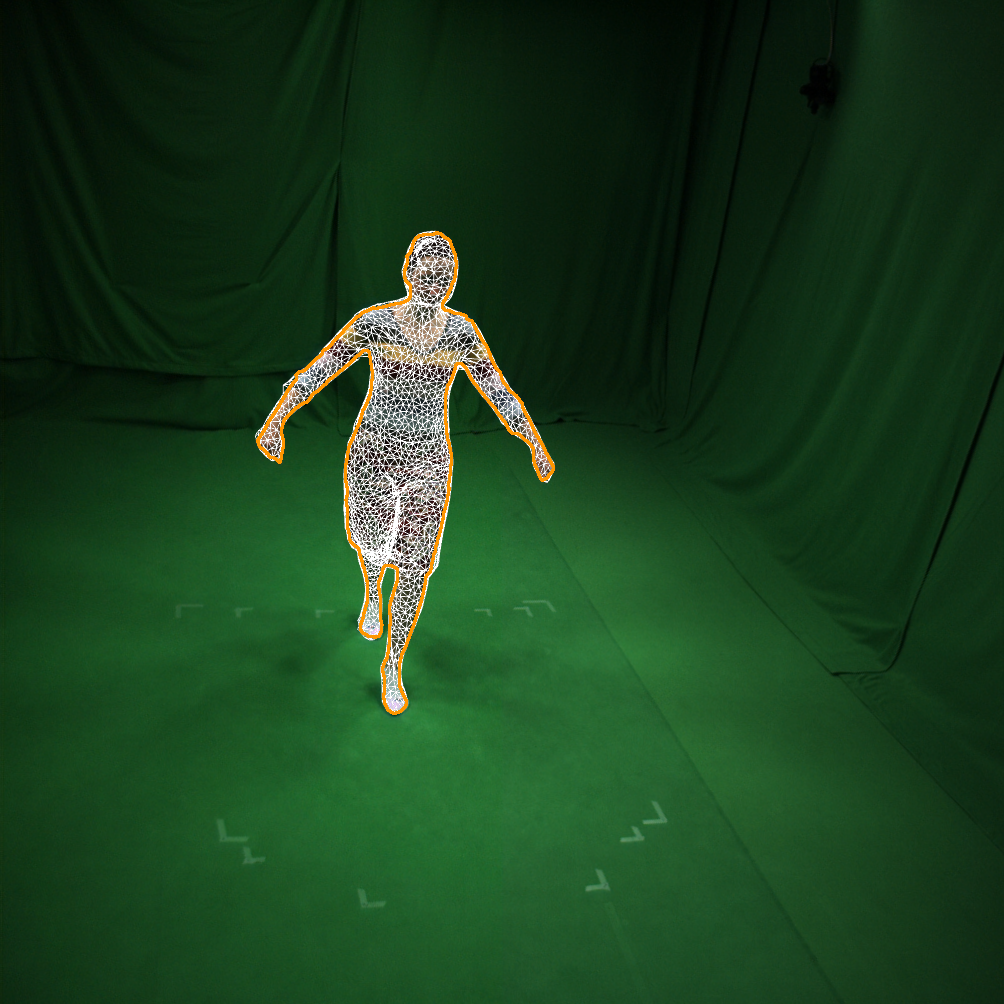}};	
\node[image,right=0pt] (F13) at (F12.east)		
{\trimmedSkirtEvaluationZoom{figures/overlap_evaluation/skirt/cam01/ours_with_silhouettes_evaluation.png}};
\node[image,below left = 0.1cm and -1.0cm of F7] (F14)	
{\trimmedSkirtEvaluationZoom{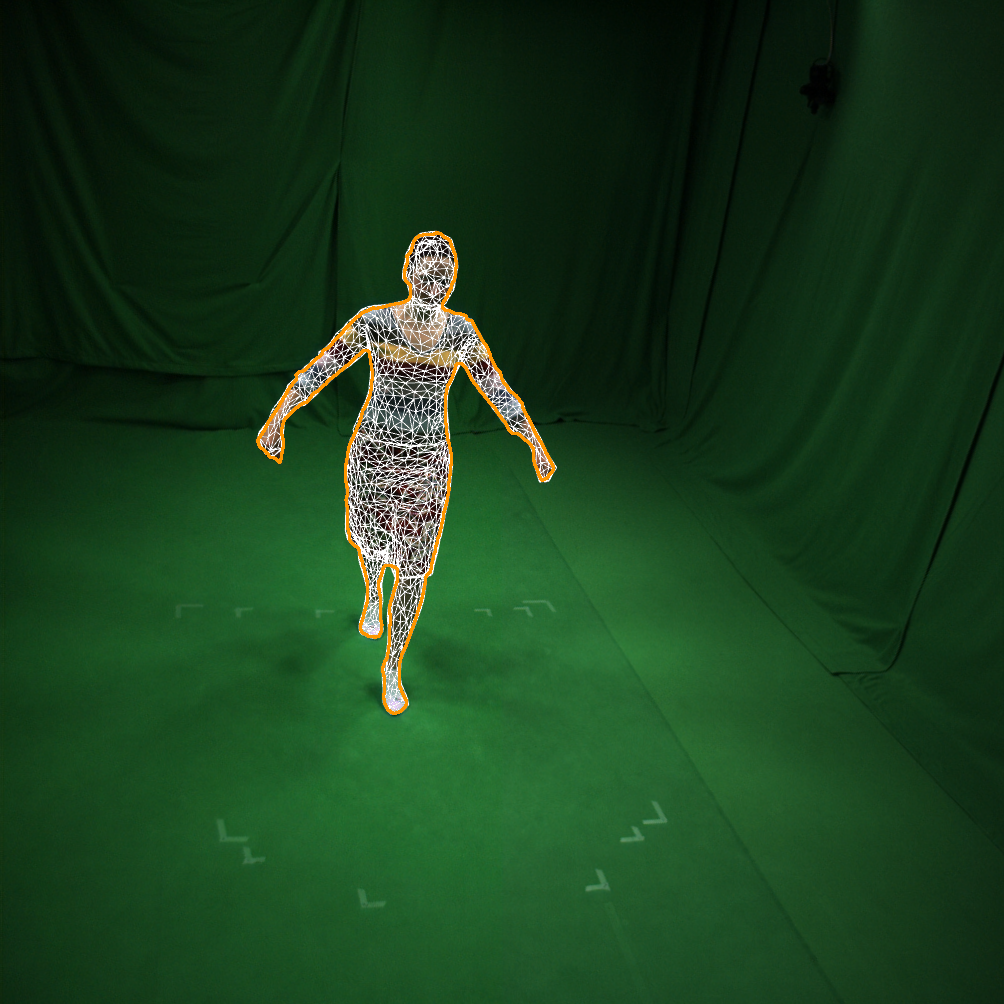}};
\node[image,right=0pt] (F15) at (F14.east)		
{\trimmedSkirtEvaluationZoom{figures/overlap_evaluation/skirt/cam01/jurguen_evaluation.png}};
\node[scale=0.8,below right = 2pt and 2pt] (F16) at (F8.south) {Stage-I};
\node[scale=0.8,right = 30pt] (F17) at (F16.east) {Stage-II};
\node[scale=0.8,right = 18pt] (F18) at (F17.east) {Stage II, with};
\node[scale=0.8,below = -4pt] (F18_) at (F18.south) {silhouettes};
\node[scale=0.8,right = 8pt] (F19) at (F18.east) {Gall \etal \cite{gall2009motion}};
\node[scale=0.8,below = -4pt] (F19_) at (F19.south) {(req. silhouettes)};
\draw[red, thick,densely dashed] (0.58,-1) rectangle (0.95,-0.1);
\end{tikzpicture}%
	\caption {Evaluation of the proposed approach in the \texttt{skirt} sequence. In orange, the ground truth contour. Stage-II significantly improves the misalignment errors present in Stage-I, caused by non-rigid deformations and skinning artifacts. Additionally, we also compare the performance of our method in ideal conditions with known silhouettes, which generate results comparable to state-of-the-art silhouette-based methods \cite{gall2009motion}. See Section \ref{sec:quantitative_evaluation} for color scheme description and further details.}%
	\label{fig:skirtEvaluation}
\end{figure}
\begin{figure}
	\centering
	\begin{tikzpicture}[image/.style={inner sep=0pt},]
	\node[image,right=0pt] (F0) at (0,0)
	{\includegraphics[trim={0cm 6.7cm 0cm 7cm}, clip,width=0.96\columnwidth]{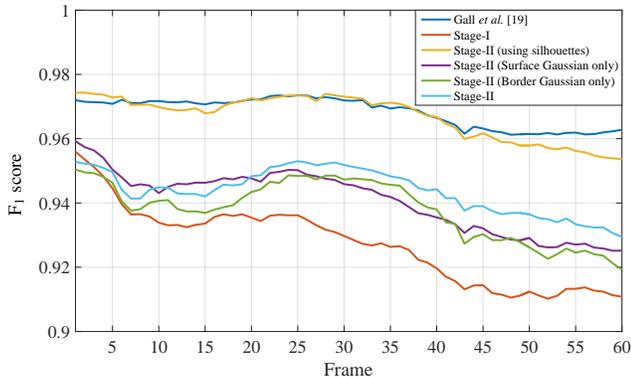}};%
	\node[rotate=90] at (-0.25,0) {\scriptsize F\textsubscript{1} score};
	\end{tikzpicture}
	\caption{Quantitative evaluation of the silhouette overlap for the \texttt{skirt} sequence. Mesh refinement using Surface Gaussians and Border Gaussians (Stage-II, in light blue) significantly improves over Stage-I, as we well as over using Surface or Border Gaussian alone. Additionally, we also show that in ideal conditions (\ie known silhouettes), our method performs comparably to the indoor silhouette-based approach of Gall \etal \cite{gall2009motion}.}%
	\label{fig:F1score}
\end{figure}

Figure \ref{fig:F1score} visualizes the F\textsubscript{1} scores across 90 frames and 8 cameras of the \texttt{skirt} sequence, for different configurations.
We evaluate each of the components of our energy, demonstrating that mesh refinement using Surface Gaussians and Border Gaussians significantly improves over Stage-I, as we well as using Surface or Border Gaussian alone.
Average F\textsubscript{1} score values when enforcing known color background ($0.9676\pm0.0056$) are comparable to the silhouette-based method from Gall \etal \cite{gall2009motion} ($0.9683\pm0.0045$).

Extensive quantitative evaluation on outdoor footage is difficult due to the lack of ground truth data, which can only be generated with laborious manual segmentation.
However, we manually segmented 10 frames of the publicly available \texttt{cathedral} dataset \cite{kim2014influence}, as well as of our new sequences \texttt{unicampus} and \texttt{pablo}.
Figure \ref{fig:overlap_evaluation} shows the silhouette overlap evaluation in these sequences, demonstrating consistent improvement after mesh refinement. Despite the challenging scenes, with uncontrolled background, our method successfully reconstructs and refines the surface of the actor model, without requiring explicit manual silhouette segmentation.
Table \ref{tab:quantitative_outdoor} presents the F\textsubscript{1} mean and standard deviation of evaluated frames in these sequences, consistently showing that our performance capture method achieves high scores even in such challenging datasets.

\newcommand{\trimmedPablo}[2][]{%
	\includegraphics[trim={5cm 1cm 27cm 3cm},clip, width=\textwidth,#1]%
	{#2}%
}
\newcommand{\trimmedHelge}[2][]{%
	\includegraphics[trim={1cm 1cm 11cm 0cm},clip, width=\textwidth,#1]%
	{#2}%
}
\newcommand{\trimmedSurrey}[2][]{%
	\includegraphics[trim={8.5cm 3.75cm 9cm 1.25cm},clip, width=\textwidth,#1]%
	{#2}%
}
\begin{figure}
	\centering
	\begin{subfigure}[b]{0.19\linewidth}
		\trimmedSurrey{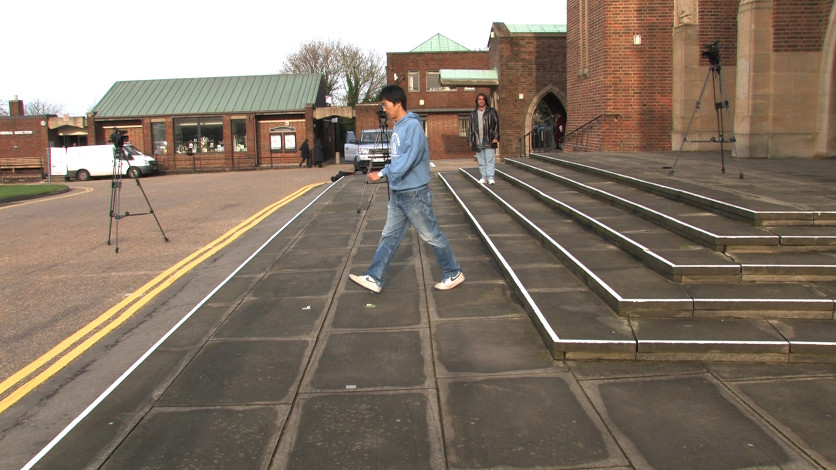}
	\end{subfigure}
	\begin{subfigure}[b]{0.19\linewidth}
		\trimmedSurrey{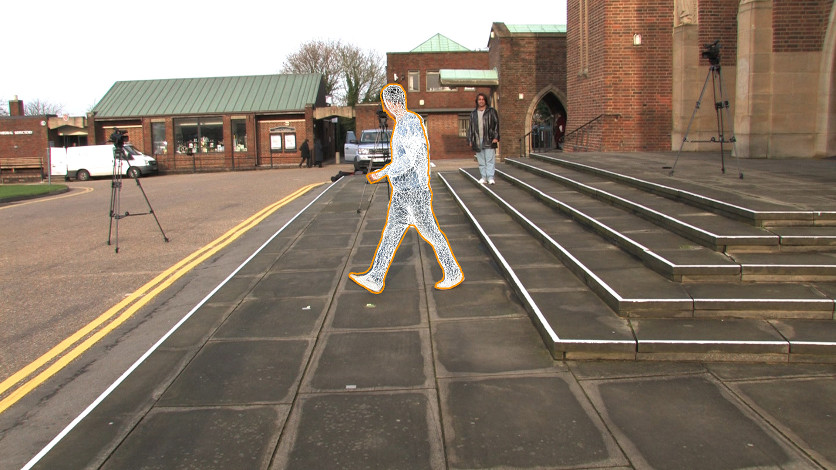}
	\end{subfigure}
	\begin{subfigure}[b]{0.19\linewidth}
		\trimmedSurrey{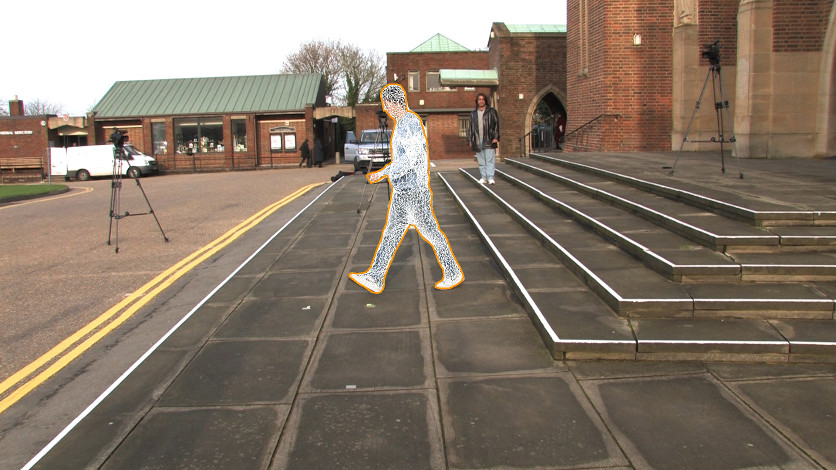}
	\end{subfigure}
	\begin{subfigure}[b]{0.19\linewidth}
		\trimmedSurrey{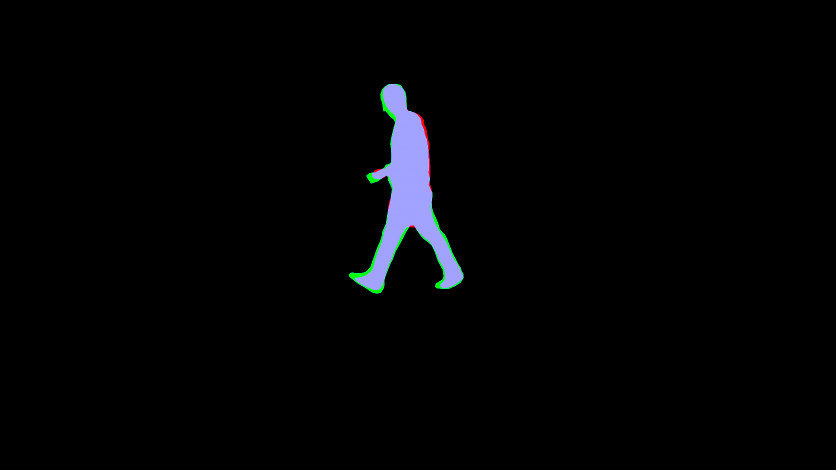}
	\end{subfigure}
	\begin{subfigure}[b]{0.19\linewidth}
		\trimmedSurrey{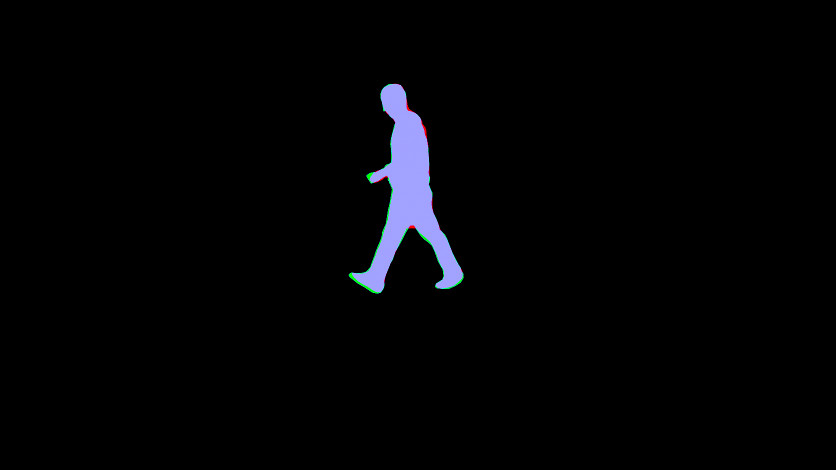}
	\end{subfigure}	
	\\[0.075cm]		
	\begin{subfigure}[b]{0.19\linewidth}
		\trimmedPablo{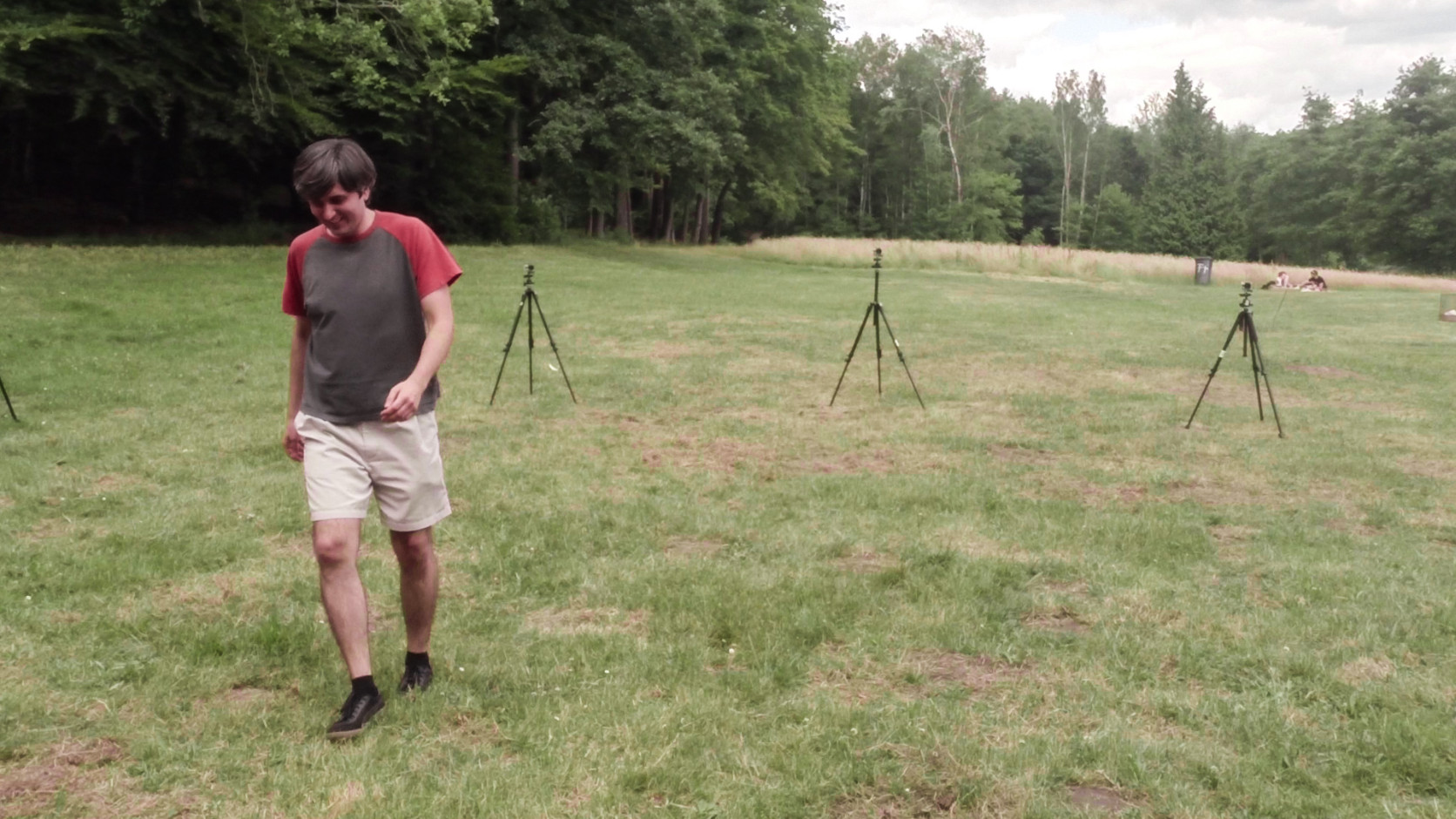}
	\end{subfigure}
	\begin{subfigure}[b]{0.19\linewidth}
		\trimmedPablo{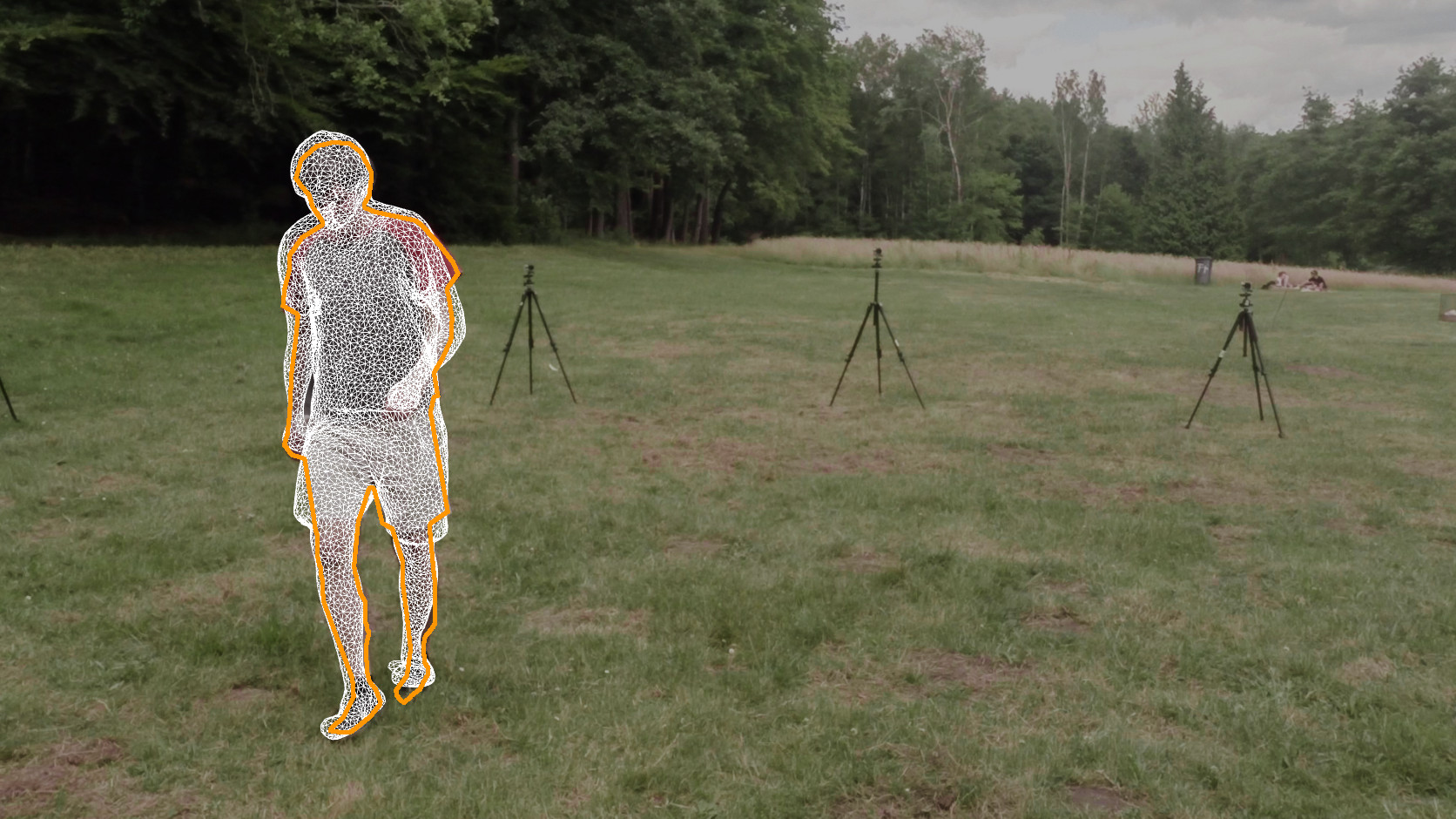}
	\end{subfigure}
	\begin{subfigure}[b]{0.19\linewidth}
		\trimmedPablo{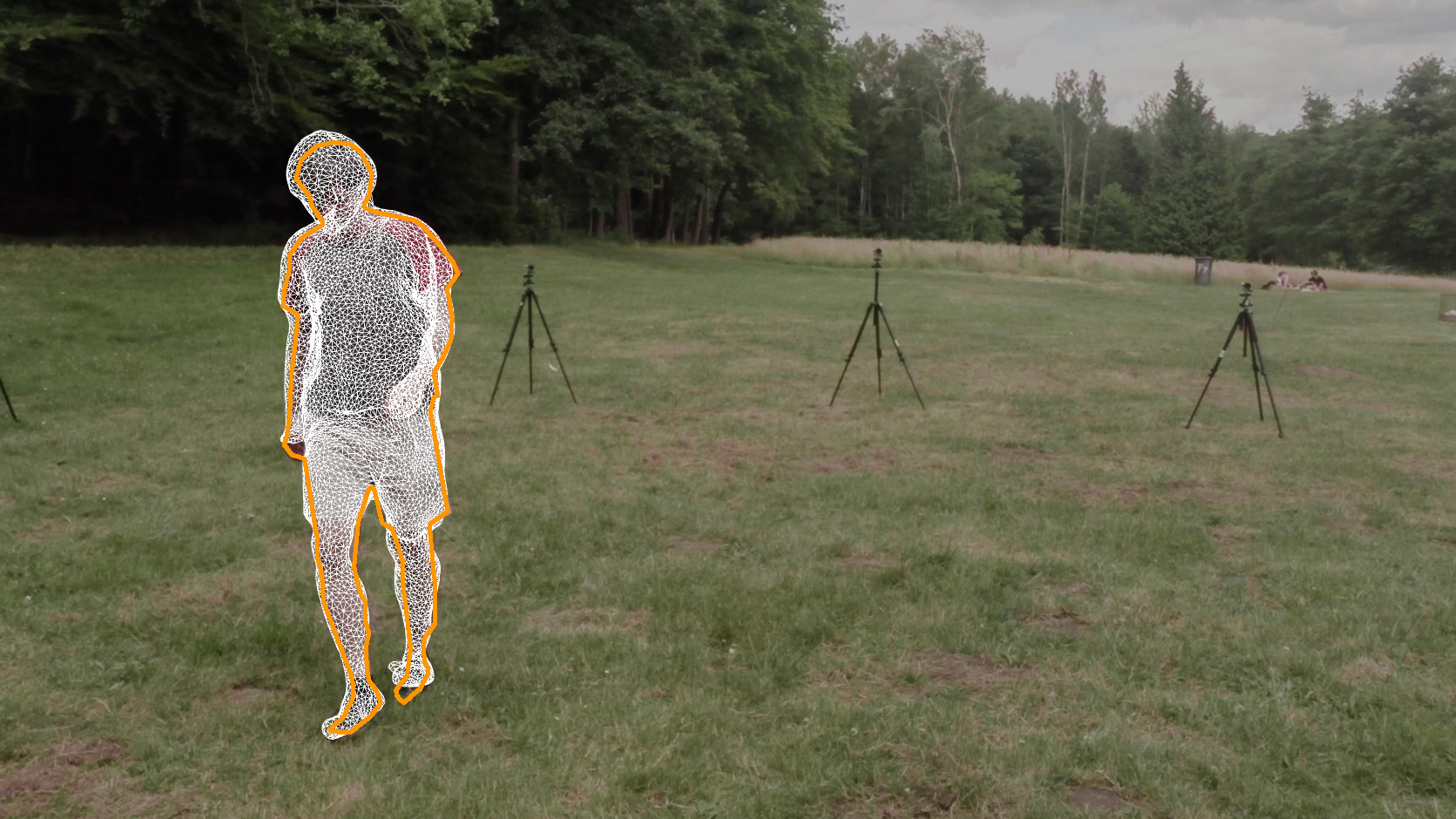}
	\end{subfigure}
	\begin{subfigure}[b]{0.19\linewidth}
		\trimmedPablo{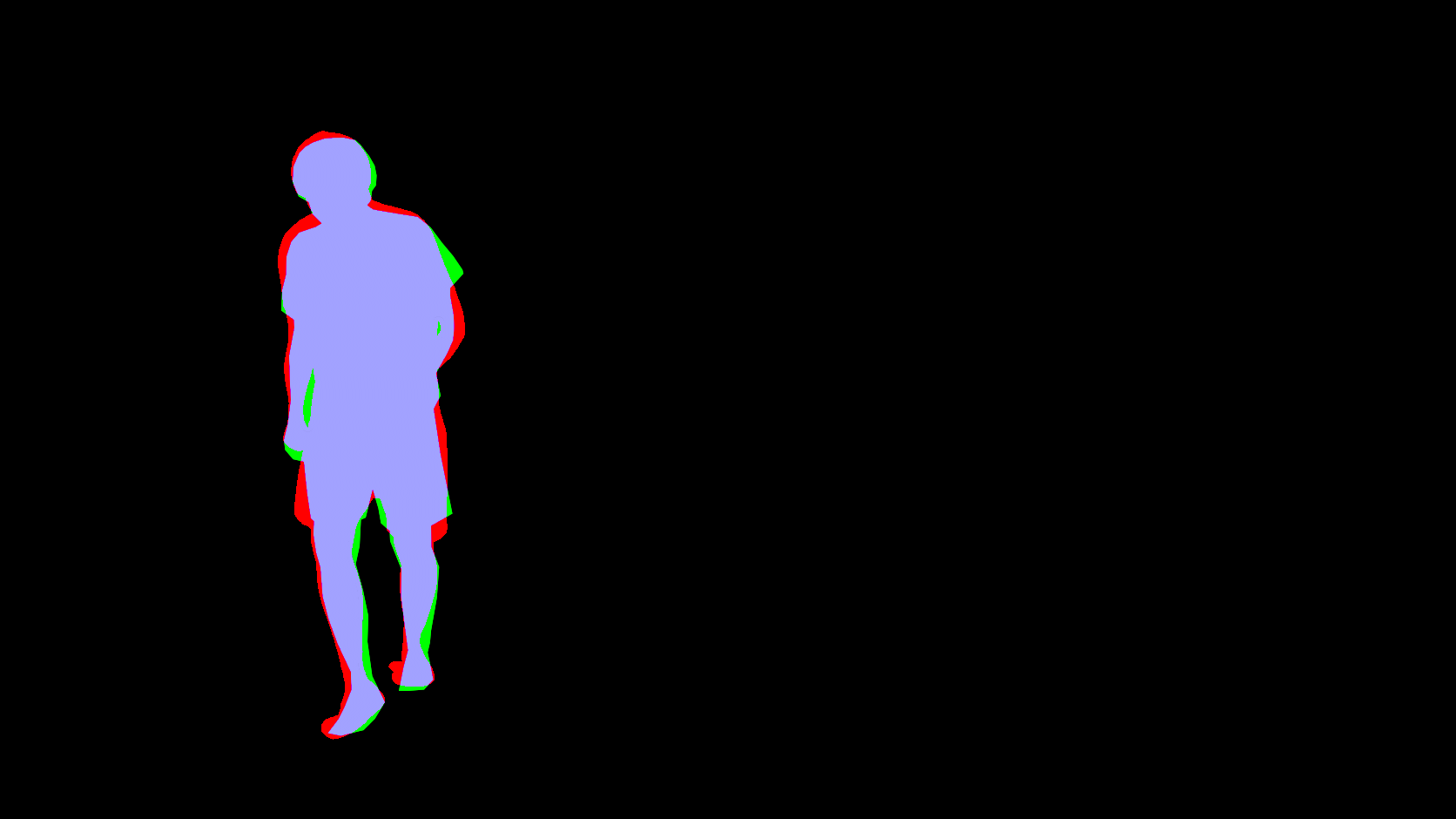}
	\end{subfigure}
	\begin{subfigure}[b]{0.19\linewidth}
		\trimmedPablo{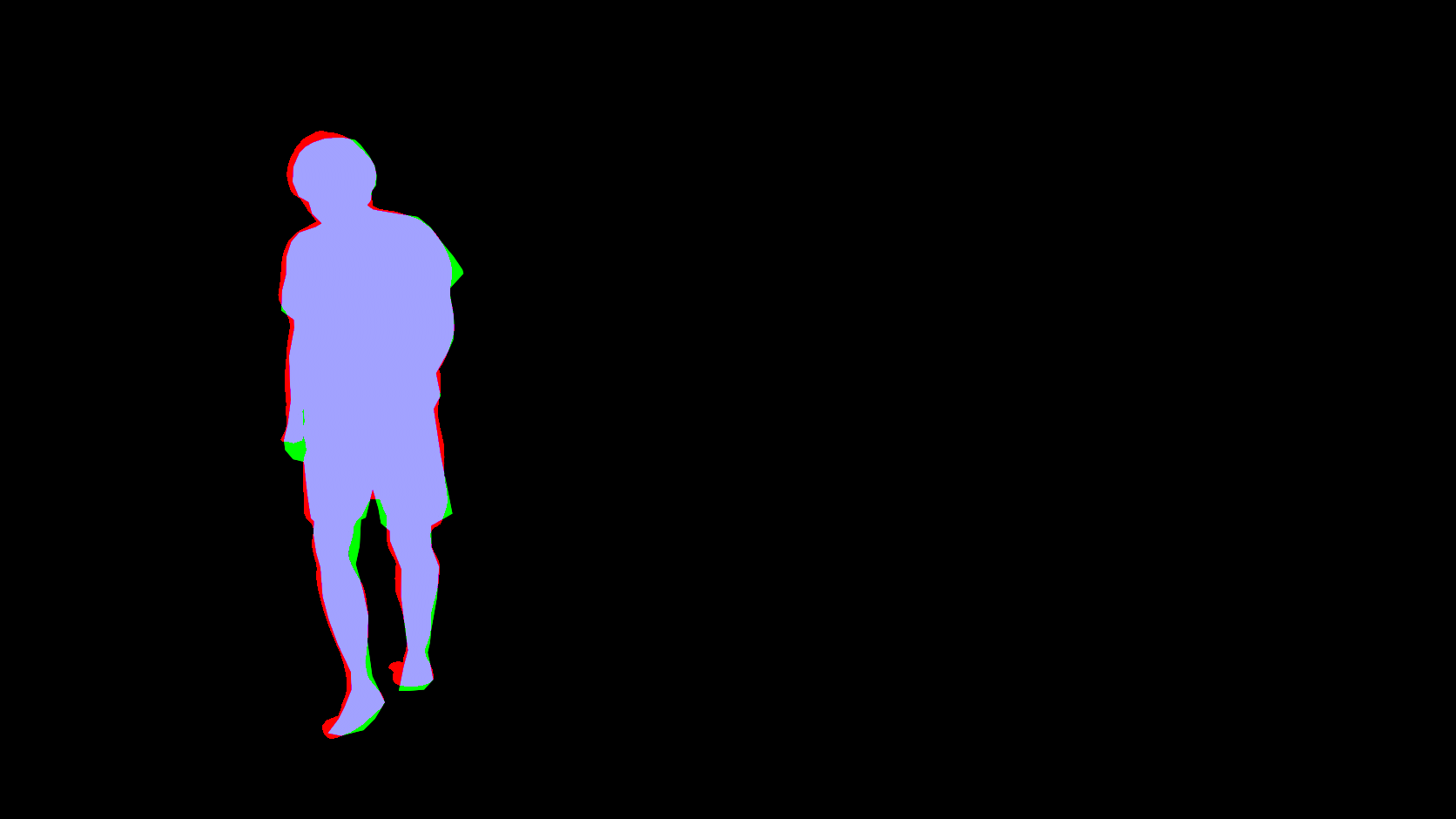}
	\end{subfigure}
	\\[0.075cm]
	\begin{subfigure}[b]{0.19\linewidth}
		\trimmedHelge{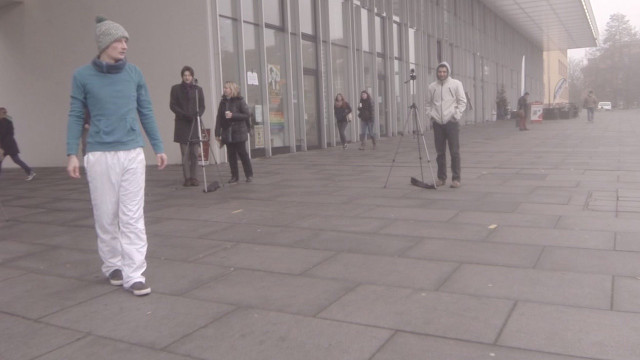}
	\end{subfigure}
	\begin{subfigure}[b]{0.19\linewidth}
		\trimmedHelge{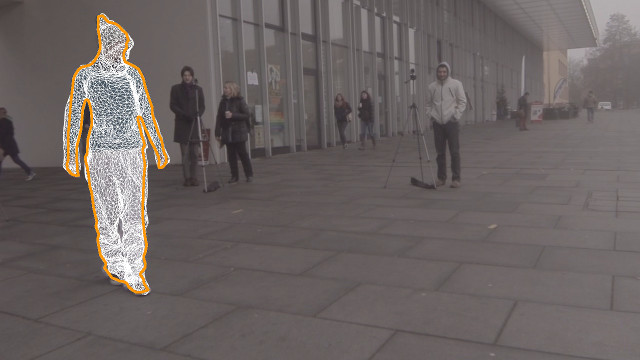}
	\end{subfigure}
	\begin{subfigure}[b]{0.19\linewidth}
		\trimmedHelge{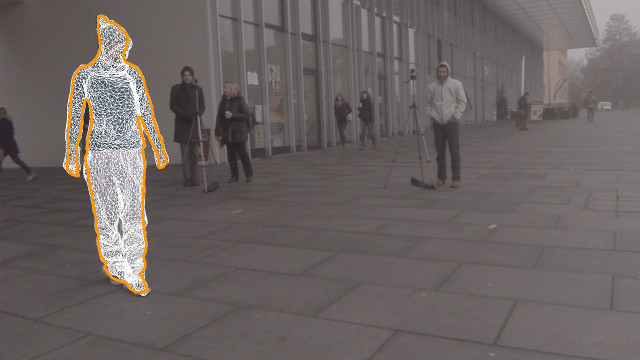}
	\end{subfigure}
	\begin{subfigure}[b]{0.19\linewidth}
		\trimmedHelge{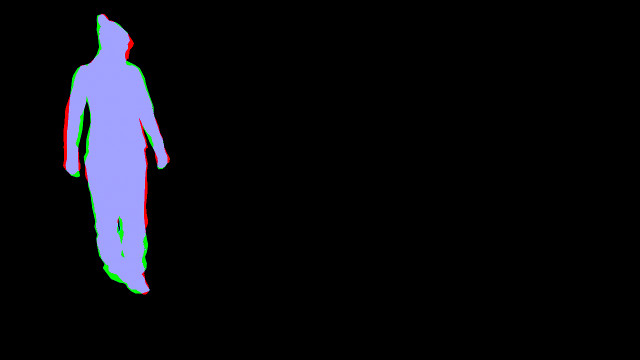}
	\end{subfigure}
	\begin{subfigure}[b]{0.19\linewidth}
		\trimmedHelge{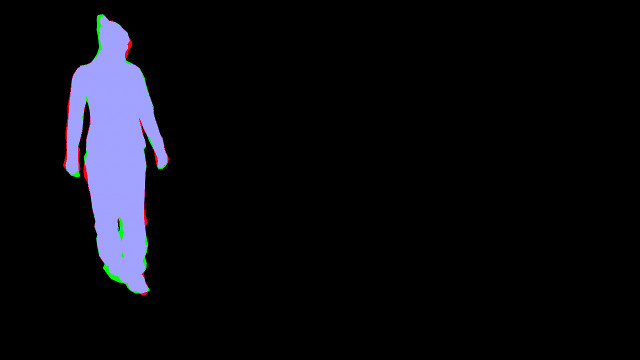}
	\end{subfigure}
	\caption{Silhouette overlap evaluation in outdoor sequences. From right to left: original frame; Stage-I mesh and ground truth contour in orange; Stage-II refined mesh and and ground truth contour in orange; Stage-I overlap silhouette evaluation; and  Stage-II refined overlap silhouette evaluation.  }\label{fig:overlap_evaluation}
\end{figure}

{
\setlength\extrarowheight{2pt}
\begin{table}
	\centering
	\begin{tabular}{ |l|c|c| } 
		\cline{2-3}
		\multicolumn{1}{c|}{} &\multicolumn{2}{c|}{F\textsubscript{1} score} \\
		\cline{2-3}
		\multicolumn{1}{c|}{}& Stage-I & Stage-II \\
		\hline
		\texttt{cathedral} & $0.9114\pm0.0077$ & $0.9362\pm0.0033$ \\ 
		\hline
		\texttt{pablo} & $0.8812\pm0.0156$ & $0.9212\pm0.0096$ \\ 
		\hline
		\texttt{unicampus} & $0.8962\pm0.0149$ & $0.9223\pm0.0083$ \\ 
		\hline
		\texttt{skirt} &  $0.9271\pm0.0122$ & $0.9676\pm0.0056$ \\ 
		\hline		
	\end{tabular}
	\caption{Quantitative evaluation of the sequences tested in this paper. The F\textsubscript{1}score of the Stage-II is consistently higher than in Stage-I.}%
	\label{tab:quantitative_outdoor}
\end{table}
}

\subsection{Qualitative Results}
\label{sec:qualitative_evaluation}
In Figures \ref{fig:teaser} and \ref{fig:results}, as well as in the supplementary video, we qualitative show reconstruction results on 4 different sequences: \texttt{skirt}, \texttt{cathedral}, \texttt{unicampus} and \texttt{pablo}.
Our results demonstrate that reconstructed meshes are temporally coherent, do not suffer from temporal noise, and maintain the level of detail of the input template without suffering from unnatural geometric deformation artifacts.
We also show textured models by reprojecting the original image frames onto the refined models, which implicitly demonstrates the accuracy of our surface reconstruction.
We believe that more advanced view-dependent texturing techniques \cite{casas20144d} would alleviate some of the remaining ghosting artifacts in the appearance --- however, we believe that ours is one of the first methods that demonstrates refined geometric and textured reconstructions of humans performing outdoors.

\newcommand{\trimmedSurreyResultsOne}[2][]{%
	\includegraphics[trim={1.25cm 2cm 3.75cm 0.8cm},clip, width=0.197\linewidth,#1]%
	{#2}%
	}
\newcommand{\trimmedSurreyResultsTwo}[2][]{%
		\includegraphics[trim={1.25cm 0.25cm 0cm 0.425cm},clip, width=0.197\linewidth,#1]%
		{#2}%
	}
\newcommand{\trimmedPabloResultsOne}[2][]{%
	\includegraphics[trim={2.5cm 1.5cm 6.5cm 3.5cm},clip, width=0.197\linewidth,#1]%
	{#2}%
}
\newcommand{\trimmedPabloResultsTwo}[2][]{%
	\includegraphics[trim={2.5cm 2.25cm 2.5cm 0.6cm},clip, width=0.197\linewidth,#1]%
	{#2}%
}
\newcommand{\trimmedSkirtResultsOne}[2][]{%
	\includegraphics[trim={2.5cm 2.0cm 2.5cm 1.5cm},clip, width=0.197\linewidth,#1]%
	{#2}%
}
\newcommand{\trimmedSkirtResultsTwo}[2][]{%
	\includegraphics[trim={1.25cm 0.5cm 1.25cm 1.5cm},clip, width=0.197\linewidth,#1]%
	{#2}%
}
\newcommand{\trimmedHelgeResultsOne}[2][]{%
	\includegraphics[trim={2.5cm 2.0cm 2.5cm 1.5cm},clip, width=0.197\linewidth,#1]%
	{#2}%
}
\newcommand{\trimmedHelgeResultsTwo}[2][]{%
	\includegraphics[trim={0cm 0.5cm 0cm 0.5cm},clip, width=0.197\linewidth,#1]%
	{#2}%
}	
	
\begin{figure*}
	\begin{tikzpicture}[image/.style={inner sep=0pt},]
		\node[image,right=0pt] (F0) at (0,0)		
		{\includegraphics[width=0.197\linewidth]{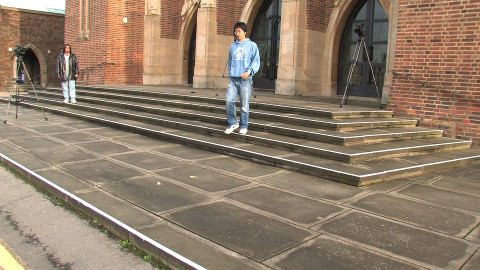}};%
		\node[image,right=2pt] (F1) at (F0.east)		
		{\trimmedSurreyResultsOne{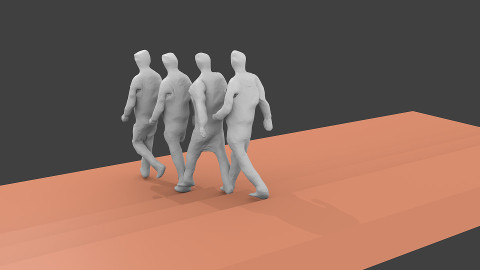}};%
		\node[image,right=2pt] (F2) at (F1.east)		
		{\trimmedSurreyResultsOne{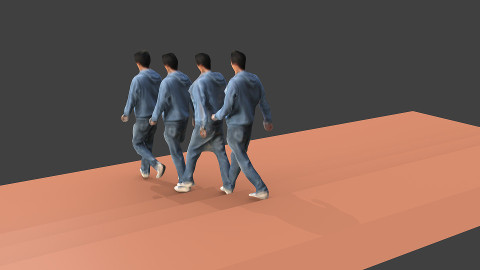}};%
		\node[image,right=2pt] (F3) at (F2.east)		
		{\trimmedSurreyResultsTwo{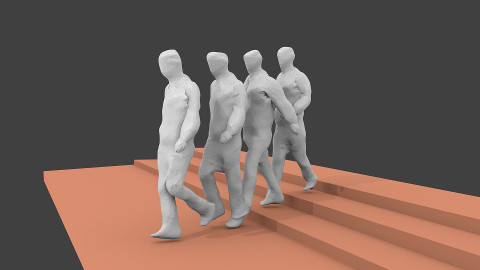}};%
		\node[image,right=2pt] (F4) at (F3.east)		
		{\trimmedSurreyResultsTwo{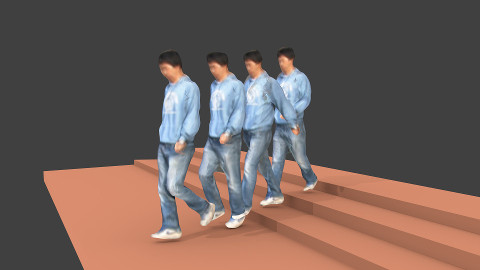}};%
		\node[image,below=2pt] (F5) at (F0.south)		
		{\includegraphics[width=0.197\linewidth]{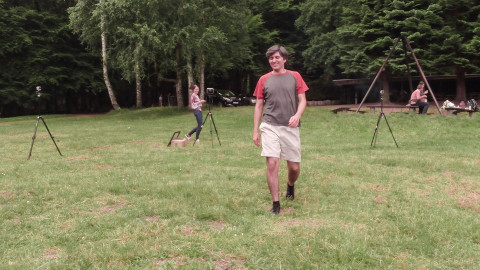}};%
		\node[image,right=2pt] (F6) at (F5.east)		
		{\trimmedPabloResultsOne{figures/results/pablo/{mesh.view00_s}.jpg}};%
		\node[image,right=2pt] (F7) at (F6.east)		
		{\trimmedPabloResultsOne{figures/results/pablo/{texture.view00_s}.jpg}};%
		\node[image,right=2pt] (F8) at (F7.east)		
		{\trimmedPabloResultsTwo{figures/results/pablo/{mesh.view02_s}.jpg}};%
		\node[image,right=2pt] (F9) at (F8.east)		
		{\trimmedPabloResultsTwo{figures/results/pablo/{texture.view02_s}.jpg}};
		\node[image,below=2pt] (F10) at (F5.south)		
		{\includegraphics[trim={0cm 1.8cm 0cm 1.4cm}, clip, width=0.197\linewidth]{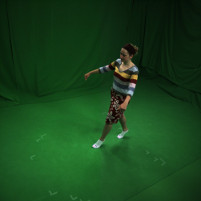}};%
		\node[image,right=2pt] (F11) at (F10.east)		
		{\trimmedSkirtResultsOne{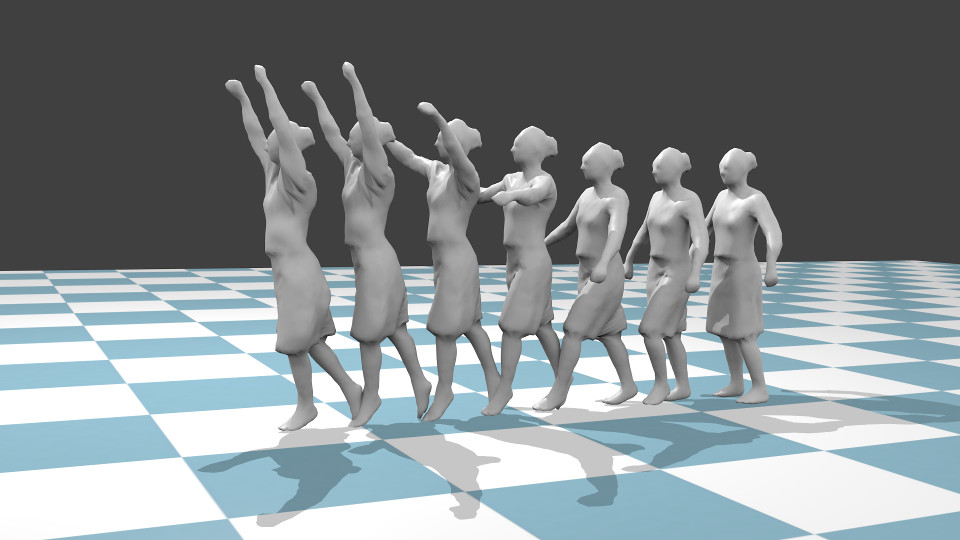}};%
		\node[image,right=2pt] (F12) at (F11.east)		
		{\trimmedSkirtResultsOne{figures/results/skirt/{texture_view01_s}.jpg}};%
		\node[image,right=2pt] (F13) at (F12.east)		
		{\trimmedSkirtResultsTwo{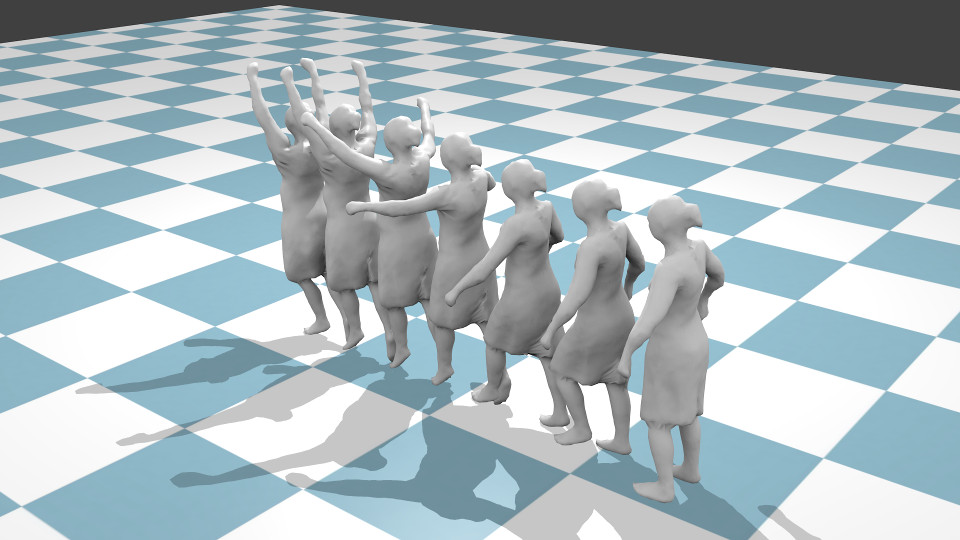}};%
		\node[image,right=2pt] (F14) at (F13.east)		
		{\trimmedSkirtResultsTwo{figures/results/skirt/{texture_view02_s}.jpg}};%
		\node[image,below=2pt] (F15) at (F10.south)		
		{\includegraphics[trim={0cm 0.55cm 1.6cm 0.5cm}, clip, width=0.197\linewidth]{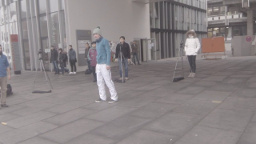}};%
		\node[image,right=2pt] (F16) at (F15.east)		
		{\trimmedHelgeResultsOne{figures/results/helge/{helge_mesh.view00_s}.jpg}};%
		\node[image,right=2pt] (F17) at (F16.east)		
		{\trimmedHelgeResultsOne{figures/results/helge/{helge_texture.view00_s}.jpg}};%
		\node[image,right=2pt] (F18) at (F17.east)		
		{\trimmedHelgeResultsTwo{figures/results/helge/{helge_mesh.view02_s}.jpg}};%
		\node[image,right=2pt] (F19) at (F18.east)		
		{\trimmedHelgeResultsTwo{figures/results/helge/{helge_texture.view02_s}.jpg}};				
	\end{tikzpicture}
	\caption {Qualitative results of our human performance capture approach. On the left, a representative input frame of each sequence. For each sequence, we show two pairs of untextured and textured reconstructions of various frames. From top to bottom: \texttt{cathedral}, \texttt{pablo}, \texttt{skirt} and \texttt{unicampus}.}
	\label{fig:results}
\end{figure*}
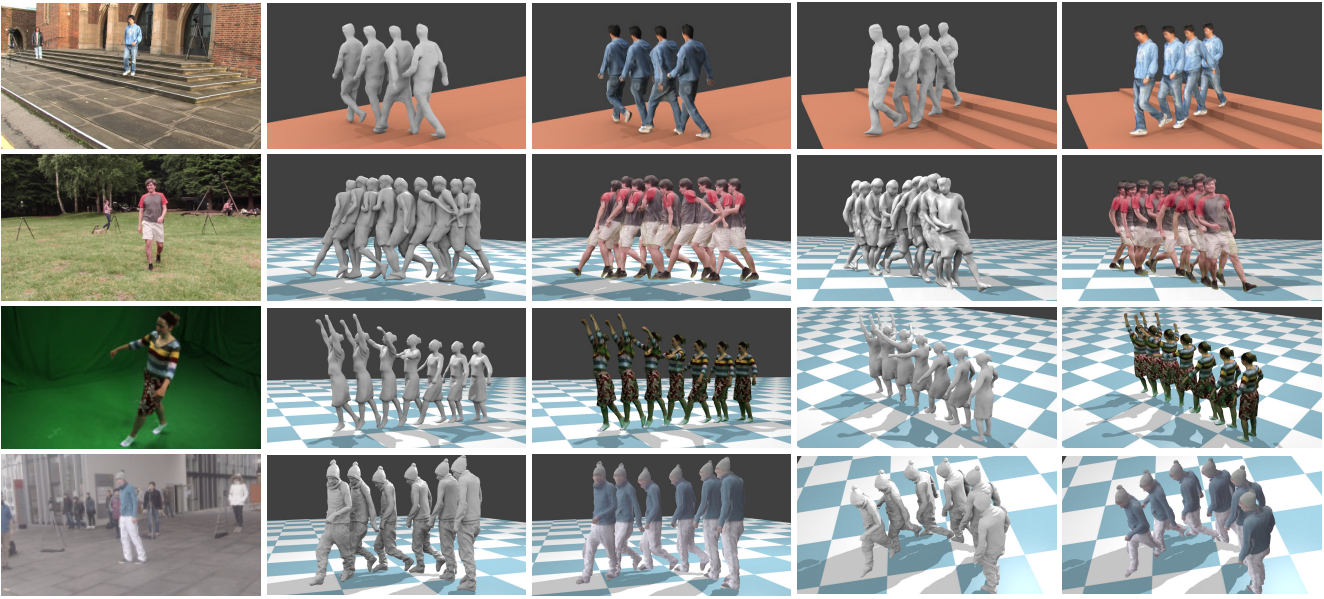


\section{Discussion and Conclusion}
We have presented one of the first model-based methods for human outdoor performance capture.
Our new unified implicit representation for both skeleton tracking and non-rigid surface refinement allows to jointly optimize pose and shape, even in scenes with unknown moving background. 
Our method fits the template to unsegmented video frames in two stages -- first skeletal pose is optimized, and subsequently both the pose and the non-rigid surface shape are refined.

While we believe our method takes a leap forward in the area of human performance capture, there are a number of challenges that remain open for future research.
Explicit illumination estimation could be incorporated into our model to better handle changes in lighting as well as time-varying shading effects happening on the surface within the same sequence.
%
Our model-based approach requires a colored template mesh for initialization and cannot cope well with complex shape or topology changes. 
Automating initialization \cite{rhodin2016eccv} and means to handle topology changes \cite{zaharescu2007transformesh} are interesting directions for future work.
Enabling the method to handle moving and unsynchronized cameras \cite{elhayek2012spatio,elhayek2015outdoor} could also be further explored, but would require a per-frame estimation of the calibration parameters.
High-frequency geometric detail could also be recovered using inverse rendering techniques -- this would potentially add even finer detail to the refined meshes.

\section*{Acknowledgements}

We thank all reviewers for their valuable feedback and The Foundry for license support.
This research was funded by the ERC Starting Grant project CapReal (335545).

{\small
\bibliographystyle{ieee}
\bibliography{robertini_3DV2016}
}
\end{document}


\title{Model-based Outdoor Performance Capture}

\author{First Author\\
Institution1\\
Institution1 address\\
{\tt\small firstauthor@i1.org}
\and
Second Author\\
Institution2\\
First line of institution2 address\\
{\tt\small secondauthor@i2.org}
}

\twocolumn[{%
	\renewcommand\twocolumn[1][]{#1}%
	\maketitle
	\begin{center}
		\centering%
		\centerline{\Large\bfseries Supplementary material}
		\vspace{3ex}
		\begin{tabular}{ |l|c|c|c|c|c|c|c| } 
			\cline{2-8}
			\multicolumn{1}{c|}{}& Type & \# Frames & Resolution & Template source & \# Vertices & $w_{\text{skin}}$ & $w_{\text{smooth}}$ \\
			\hline
			\texttt{cathedral} & outdoor &  65 & 1920x1080&   visual hull &4473 & 0.001 & 0.001\\ 
			\hline
			\texttt{pablo} &outdoor & 200 & 1920x1080 & laser scan & 9161  & 0.001 & 0.045 \\ 
			\hline
			\texttt{unicampus} & outdoor & 133  & 1280x720 & laser scan & 6155 & 0.001 & 0.025 \\ 
			\hline
			\texttt{skirt} &  indoor & 200 & 1004x1004 & laser scan& 3862 & 0.001 & 0.005 \\ 
			\hline		
		\end{tabular}
		\captionof{figure}{Technical details of the sequences used in this paper}
		\label{tab:sequences}
	\end{center}%
}]

\section{Sequences}
Table \ref{tab:sequences} presents the details of the sequences used in this paper. Note that we do not only demonstrate results in sequences that use a detailed laser scan as a input template, but also more challenging scenarios where the scan was not available (\ie the \texttt{cathedral} sequence).
In this situations, we manually segmented one frame of the sequence and performed silhouette-based reconstruction to obtained a static reconstruction, which we use as a template.